\documentclass{article}

\usepackage{microtype}
\usepackage{graphicx}
\usepackage{subfigure}
\usepackage{booktabs} 

\usepackage{hyperref}

\usepackage{pifont}
\newcommand{\cmark}{\ding{51}}%
\newcommand{\xmark}{\ding{55}}%

\usepackage[accepted]{icml2024}

\usepackage{amsmath}
\usepackage{amssymb}
\usepackage{mathtools}
\usepackage{amsthm}

\usepackage{CJKutf8}

\usepackage{soul}
\newcommand{\hlc}[2][yellow]{{\sethlcolor{#1} \hl{#2}}}

\newlength{\exampleWidth}
\usepackage{float}

\usepackage[capitalize,noabbrev]{cleveref}

\usepackage{tcolorbox}

\renewcommand{\algorithmicinput}{\textbf{Given:}}
\hypersetup{colorlinks=true,urlcolor={black!40}}

\theoremstyle{plain}

\theoremstyle{definition}

\theoremstyle{remark}

\usepackage[textsize=tiny]{todonotes}

\usepackage{tabularx}
\newcolumntype{L}{>{\raggedright\arraybackslash}X}

\usepackage{enumitem}

\icmltitlerunning{}

\begin{document}

\twocolumn[
\icmltitle{From Jack of All Trades to Master of One: \\Specializing LLM-based Autoraters to a Test Set}



\icmlsetsymbol{equal}{*}

\begin{icmlauthorlist}
\icmlauthor{Mara Finkelstein}{}
\icmlauthor{Daniel Deutsch}{}
\icmlauthor{Parker Riley}{}
\icmlauthor{Juraj Juraska}{}
\icmlauthor{Geza Kovacs}{}
\icmlauthor{Markus Freitag}{}
\end{icmlauthorlist}
\centering
\texttt{\{marafin,dandeutsch,prkriley,jjuraska,geza,freitag\}@google.com}


\icmlcorrespondingauthor{Mara Finkelstein}{marafin@google.com}

\icmlkeywords{Machine Learning, ICML}

\vskip 0.3in
]




\begin{abstract}
As LLMs continue to become more powerful and versatile, human evaluation has quickly become intractable at scale and reliance on automatic metrics has become the norm. Recently, it has been shown that LLMs are themselves state-of-the-art evaluators for many tasks. These \textit{Autoraters} are typically designed so that they generalize to new systems \textit{and} test sets. In practice, however, evaluation is performed on a small set of fixed, canonical test sets, which are carefully curated to measure certain capabilities of interest and are not changed frequently. In this work, we design a method which specializes a prompted Autorater to a given test set, by leveraging historical ratings on the test set to construct in-context learning (ICL) examples. We evaluate our \textit{Specialist} method on the task of fine-grained machine translation evaluation, and show that it dramatically outperforms the state-of-the-art XCOMET metric by 54\% and 119\% on the WMT'23 and WMT'24 test sets, respectively. We perform extensive analyses to understand the representations learned by our Specialist metrics, and how variability in rater behavior affects their performance. We also verify the generalizability and robustness of our Specialist method for designing automatic metrics across different numbers of ICL examples, LLM backbones, systems to evaluate, and evaluation tasks.
\end{abstract}

\section{Introduction}
\label{sec:intro}

While evaluation of natural language generation (NLG) systems has been a long-standing challenge, its importance has come to the fore in the era of large language models (LLMs). Moreover, while human evaluation has historically been considered the gold standard for measuring model quality, it has become a key bottleneck during model development. In addition to being costly, slow, and difficult to scale, human evaluation is also limited by subjectivity \citep{krishna2023longeval} and high variability in judgments across human raters \citep{karpinska2021perils,riley2024finding,zhang2024diverging}, even for a fixed example in a given test set. Increasingly, automatic metrics are replacing human evaluation for measuring the quality of generative models, and LLMs themselves have been shown to be state-of-the-art evaluators across a range of capabilities \citep{kim2023prometheus, kim2024prometheus, vu2024foundational, li2023generative}. Automatic metrics have also become a key component of LLM training, in which they are used as reward models during reinforcement learning-based preference optimization \citep[e.g. RLHF;][]{ouyang2022training}.

These ``LLM-as-a-Judge'' evaluators are often referred to as \textit{Autoraters} \citep{vu2024foundational}. Some Autoraters are finetuned on human judgements \citep{kim2023prometheus, kim2024prometheus, vu2024foundational, li2023generative}, while others are simply prompted \citep{kocmi2023gemba}, with a few human judgements provided as demonstrations. Prompting LLMs with in-context learning (ICL) examples (also known as demonstrations) is a common approach for eliciting their reasoning and instruction-following capabilities \citep{tanzer2023benchmark, yan2023understanding}. While traditional automatic evaluation metrics (and reward models) predict scalar quality scores, the transition towards generative Autoraters for evaluation opens up the possibility to elicit feedback more flexibly, including fine-grained and interpretable feedback \citep{fernandes2023devil}.

The race to build ever-more-performant LLMs has accelerated not only this shift from human to automatic evaluation, but has also brought demand for standard test sets on which LLM quality is measured and compared \citep{hendrycks2020measuring, liang2022holistic, zheng2023lmsys, zheng2023judging}. Evaluating new systems on a fixed set of benchmarks, which are carefully curated to measure certain capabilities of interest and are not changed frequently, allows for fair comparison against previous work and is the standard in the literature. Thus, while automatic metrics are typically designed so that they generalize to new systems \textit{and} test sets, in practice, it is very important that the evaluation metric being used work well across systems on the given test set, and less important that the metric generalize to other, unseen and unused test sets. In this work, we propose a simple and highly effective method to build LLM-based Autoraters which are specialized to a given test set, by leveraging historical ratings on the test set to construct ICL examples.

\begin{figure*}[t]
\centering
\includegraphics[scale=0.17]{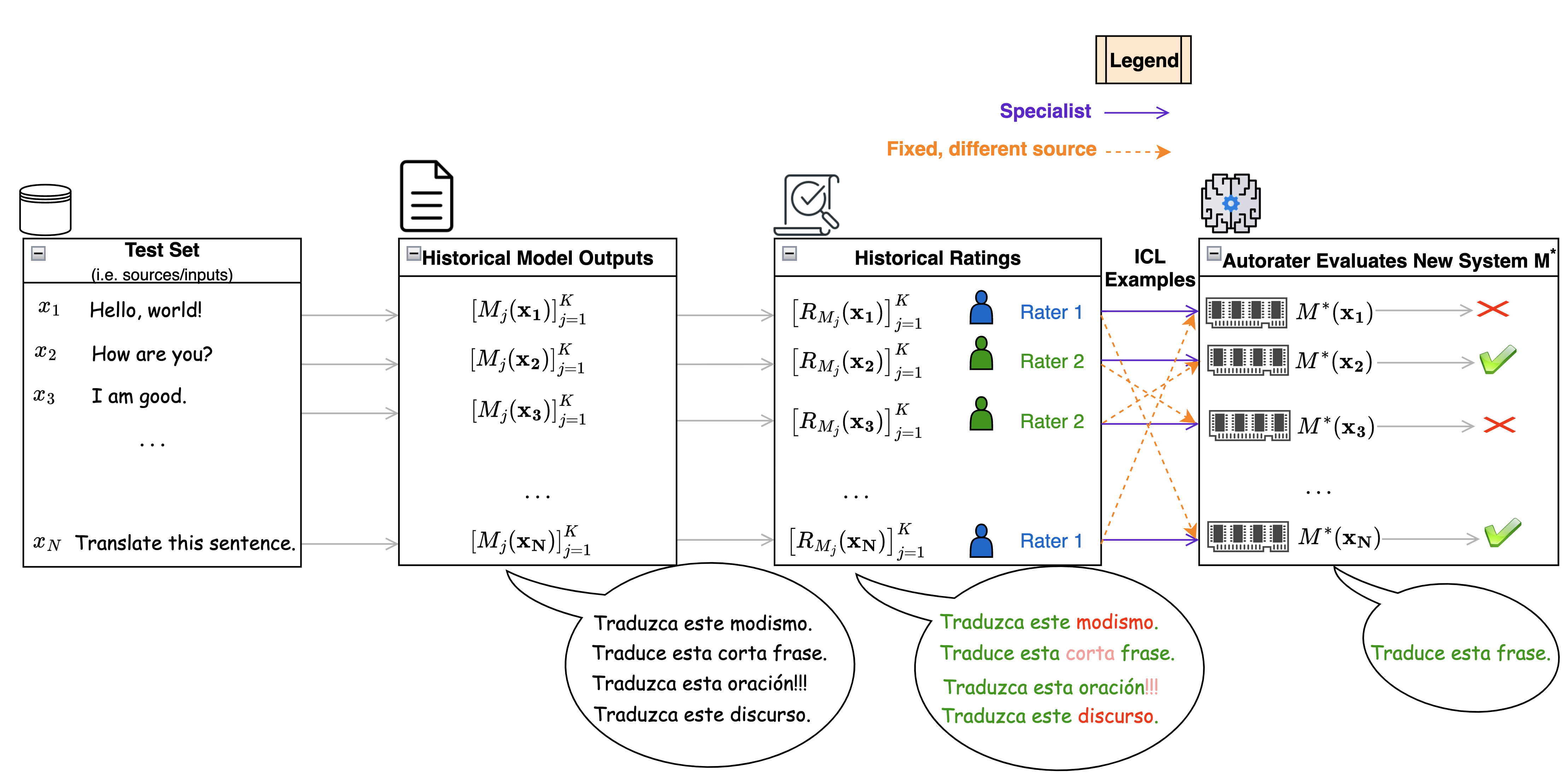}
\caption{Illustration of the \texttt{Specialist} method, compared against the \texttt{Fixed}, \texttt{different} \texttt{source} baseline, for prompting an LLM-based Autorater. Both methods (i) construct a unique set of demonstrations (i.e., ICL examples) for every test set example, consisting of historical ratings from different system outputs for some \textit{fixed source}, and (ii) provide demonstrations from the \textit{same rater} as the test rating ground truth. The difference between these methods is that the \texttt{Specialist} ICL examples consist of ratings of outputs from the \textit{same source} as the test example.}
\label{fig:specialist_diagram}
\end{figure*}

Our contributions can be summarized as follows:
\begin{itemize}
    \item We propose a novel method for constructing LLM-based Autoraters for NLG evaluation, which are specialized to a given test set. This method only requires multi-shot prompting (no finetuning).
    \item We show that this method can be used to construct a state-of-the-art automatic metric for fine-grained machine translation (MT) evaluation, which we call Specialist AutoMQM. This metric dramatically outperforms the existing state-of-the-art, achieving character-level F1 improvements of 54\% and 119\% on the WMT'23 and WMT'24 test sets, respectively, relative to XCOMET \citep{guerreiro2023xcomet}.
    \item We perform extensive ablations and analyses to verify that the representations learned from the ICL examples by our Specialist AutoMQM model are non-trivial and robust. For example, we show that Specialist AutoMQM performance scales with number of ICL examples used, but that this improvement in performance cannot simply be attributed to copying errors from ICL examples, and that ICL examples also teach the model which errors not to predict.
    \item We investigate how variability in judgments across different human raters affects performance of Specialist AutoMQM, and conclude that this metric specializes not only to the test set, but also to the rater.
    \item We show that the Specialist method is robust to the choice of LLM and to the systems being evaluated, and also that this method generalizes to the different, but related, task of score prediction for machine translation.
\end{itemize}
\section{Related Work}
\label{sec:related-work}

\paragraph{LLM-as-a-Judge Autoraters} With the need to evaluate LLMs across an ever-increasing range of capabilities, human evaluation has quickly become intractable at scale and reliance on automatic metrics has become the norm. Recently, it has been shown that, for many tasks, LLMs are themselves state-of-the-art evaluators \citep{kim2023prometheus, kim2024prometheus, vu2024foundational, li2023generative}. The emerging paradigm for development of these LLM-as-a-Judge Autoraters takes advantage of their generative architecture to flexibly perform evaluation across a range of tasks and protocols, by simply specifying the evaluation rubric in the Autorater prompt. The FLAMe \citep{vu2024foundational}, Auto-J \citep{li2023generative}, and Prometheus \citep{kim2023prometheus, kim2024prometheus} Autoraters are all finetuned on human (and, for Auto-J, LLM-generated) preference data covering a wide range of evaluation tasks, including both direct assessment and pairwise ranking. Generative Autoraters also facilitate moving from traditional score-based evaluation to more flexible, fine-grained, and interpretable evaluation protocols \citep{fernandes2023devil, kocmi2023gemba}. However, \citet{kamoi2024evaluating} showed that GPT-4 and Claude-3 have low recall in detecting errors made by LLMs, and explanations from LLM-based error detectors are unreliable. Specialization of LLM-as-a-Judge Autoraters to a given test set has not been previously explored, however, and we show that our \textit{Specialist} method works for Autoraters across both score-based and fine-grained evaluation tasks.

\paragraph{Modeling Rater Behavior}
For many (NLG) tasks on which LLMs are evaluated, there is high variability in judgments across human raters \citep{karpinska2021perils, riley2024finding}, even for a fixed example in a given test set. For some tasks (such as open-ended text generation tasks), the evaluation criteria have some degree of subjectivity \citep{krishna2023longeval}. Especially for expert-level evaluation tasks, differences in rater quality and conscientiousness can manifest as inter-annotator disagreement \citep{karpinska2021perils}. Raters can also have different stylistic preferences, and some grade more leniently or harshly than others \citep{riley2024finding}. Rater training and precise annotation guidelines can alleviate these differences to some extent, but are not guaranteed to eliminate them. Given these differences in rater behavior, several recent studies have sought to model behavior of multiple raters when designing automatic metrics \citep{zhang2024diverging, geva2019we, chen2024seeing, golazizian2024cost}. Many of these studies seek to predict preferences across different raters, or to model the rater preference distribution. However, this remains a poorly understood problem, and is complicated by differences in rater behavior across evaluation tasks and setups. In this work, rather than attempting to model behavior of multiple raters, we investigate whether our proposed method effectively specializes to a single rater. In practice, modeling a single, high-quality rater is often more desirable than modeling multiple, noisy raters.

\paragraph{Machine Translation Evaluation} In this work, we focus on the task of machine translation (MT) evaluation to investigate the effectiveness of our \textit{Specialist} method for constructing automatic evaluation metrics. MT is a core NLG task, and automatic MT evaluation is one of the most well-studied evaluation problems in Natural Language Processing (NLP; \citet{callison2008proceedings, freitag2023results}). Traditional model-based automatic MT evaluation metrics are regression-based neural networks (typically encoder-only or encoder-decoder models) finetuned on human judgements of translation quality to predict scalar scores \citep{sellam2020bleurt, rei2020comet, juraska2024metricx}.

In line with broader trends, research in automatic MT evaluation has recently shifted towards the LLM-as-a-Judge Autorater paradigm, with promising results both in terms of evaluator performance (correlation with human judgements), as well as interpretability. \citet{kocmi2023large} showed that LLMs prompted to predict scalar quality scores are state-of-the-art evaluators of MT quality at the system level (though still lag behind traditional finetuned MT evaluation metrics at the segment level).

While these score-based metrics have high correlation with human judgments, the scores produced by these metrics are difficult to interpret, and do not provide any actionable insights into the limitations of the model being evaluated or how to improve it \citep{xu2024llmrefine,zhang2024learningothersmistakesfinetuning}. Recent work in creating interpretable \textit{automatic} MT metrics has built upon an existing, state-of-the-art framework for interpretable \textit{human} evaluation of translation quality: the Multidimensional Quality Metrics \citep[MQM;][]{lommel2014multidimensional, freitag2021experts} framework, in which professional annotators are asked to identify and label individual error spans in MT outputs, along with the corresponding error category (e.g., \textit{fluency}, \textit{accuracy}, etc.) and severity (\textit{minor}, \textit{major}, or \textit{critical}). \citet{fernandes2023devil} and \citet{kocmi2023gemba} showed that LLMs can be few-shot prompted to provide fine-grained MQM error annotations of MT outputs. However, these prompted, LLM-based Autoraters still underperform \citep{freitag2023results} XCOMET \citep{guerreiro2023xcomet}, an encoder model finetuned on human-annotated MQM data, which predicts both scalar quality scores (with a regression head) and error spans (non-generatively) via token-level tagging.

In this work, we refer to any automatic MT metric which mimics (human expert-based) MQM as \textit{AutoMQM}. That is, an AutoMQM metric predicts error spans and, optionally, identifies error categories and/or severities according to the MQM framework. This term does not refer to any specific model, but rather to the type of metric. In the remainder of this work, we focus on building and evaluating a state-of-the-art AutoMQM metric for MT evaluation, given that next-generation NLG evaluation is increasingly focused on interpretability. As a baseline, we also evaluate our proposed \textit{Specialist} method on the task of MT evaluation via scalar score prediction in \S\ref{ablation:scorer}.
\section{\textit{Specialist} Method}
\label{sec:method}

In this work, we propose the \textit{Specialist} method for development of a prompted LLM-as-a-Judge metric, which specializes the metric to a given test set based on the ICL examples provided. This method will be phrased in terms of the machine translation (MT) evaluation task, but its formulation generalizes to any natural language generation (NLG) evaluation task (i.e., any task which evaluates output from generative models).

\paragraph{Prerequisites} First, we establish some basic terminology. The objective is to evaluate the performance of an MT system (i.e., model) $M$ on a fixed \textit{test set}. A test set simply consists of a set of \textit{sources} $X$, which are the inputs to the system(s) to be evaluated. (In this setting, we do \textit{not} require access to gold reference translations of these sources.) The \textit{translations} $Y_{M}$ are the outputs of $M$ on the test set: that is, $Y_{M} = \left\{M(x): x \in X\right\}$. Evaluation of system $M$ on the test set (whether by MQM, AutoMQM, score prediction, etc.) produces a set of \textit{ratings} $R_{M} = \left\{\text{rating}(y): y \in Y_{M} \right\}$ for the translations $Y_{M}$.

\paragraph{Specialist Algorithm} Informally, the Specialist method can be summarized as follows: Given access to a test set $X$ augmented with historical translation quality ratings from multiple systems, and given the predictions of a new translation system $M^{*}$ on this test set, the Specialist metric evaluates the quality of system $M^{*}$ on $X$ by prompting an LLM as follows: For every example (consisting of a translation $M^{*}(x)$ of some input $x \in X$), construct ICL examples for this test set example from all ratings of historical translations of the same input $x$. In this work, we primarily consider the \textit{pseudo-SxS} setting where, for each example, all historical ratings were performed by a fixed human rater. See Figure \ref{fig:specialist_diagram} for an illustration of the Specialist method.

\definecolor{light}{gray}{0.7}

\begin{algorithm}
  \caption{Specialist Method for Automatic Evaluation}
  \label{alg:specialist}
\begin{algorithmic}[1]
  \INPUT
  \STATE {Test set $X = \left\{x_i\right\}_{i=1}^{K}$}
  \STATE {Translation system $M^{*}$ to evaluate}
  \REQUIRE
  \STATE {Off-the-shelf LLM $E$ to use as the prompted evaluator}
  \STATE {Set $R$ of ratings on $X$ for $N$ translation systems: $R = \left\{R_{M_{j}}\right\}_{j=1}^{N}$, where $M_j \neq M^{*}$ for all $j \in \{1, \dots, N\}$}. \\ \textcolor{gray}{Pseudo-SxS Constraint: For each $i$, ratings $\left\{R_{M_j}[i]\right\}_{j=1}^{N}$ were performed by a single rater.}
  \ENSURE Ratings $R_{M^{*}}$ of system $M^{*}$ on test set $X$
  \STATE {$R_{M^{*}} \gets []$}
  \STATE {\textcolor{gray}{\# Iterate over examples in the test set}}
  \FOR{$i \gets 1$ to $K$}
  \STATE {\textcolor{gray}{\# Construct ICL examples from all historical ratings of the same input $x_i$}}
  \STATE {$\left(\texttt{ICL examples}\right)_i = \left\{R_{M_j}[i]\right\}_{j=1}^{N}$}
  \STATE {\textcolor{gray}{\# Compute output of model $M^*$ on this test set example}}
  \STATE {$Y_{M^{*}}^{i} = M^{*}(x_i)$}
  \STATE {\textcolor{gray}{\# Prompt the LLM $E$ to evaluate the translation of the new system $M^{*}$ on the same input $x_i$, given the historical ratings}}
  \STATE {$R_{M^{*}}^{i} = E\Big(\left(\texttt{ICL examples}\right)_i, x_i, Y_{M^{*}}^{i}\Big)$}
  \STATE {$\texttt{Append } R_{M^{*}}^{i} \texttt{ to } R_{M^{*}}$}
  \ENDFOR
  \STATE {$\textbf{Return } R_{M^{*}}$}
\end{algorithmic}
\end{algorithm}

More formally, Algorithm \ref{alg:specialist} outlines the details of our proposed method for specializing an automatic evaluation metric to a test set. This method requires access to multiple (historical) sets of (human-generated) ratings $R_{M_j}$ (from different translation systems $M_j$) on the same test set. The \textit{pseudo-SxS} setting primarily considered in this work has the additional requirement that, for each test set example $x_i$, all ratings $\left\{R_{M_j}[i]\right\}_{j=1}^{N}$ were performed by a fixed rater. Note that, in the setting we explore here, the different translations for each input example come from different translation systems, but they could in principle also be sampled from a single model (e.g., using a diversity-promoting sampling algorithm).

The evaluation metric itself is a prompted LLM, and the Specialist method constructs ICL examples (i.e., demonstrations) to be used for prompting on a per-example basis, so that ICL examples are unique for every example in the test set. In particular, for a given input $x_i$ in the test set, the ICL examples are constructed from all of the (historical) ratings of this same example (line 9 in Algorithm \ref{alg:specialist}). That is, given a new translation system $M^{*}$ to evaluate on the test set, the ICL examples used for evaluation of the translation $Y_{M^{*}}^{i} = M^{*}(x_i)$ are given by $\left\{R_{M_j}[i]\right\}_{j=1}^N$. Once the ICL examples are constructed, the LLM is prompted with these demonstrations, as well as the corresponding source $x_i$ and model translation $Y_{M^{*}}^{i}$ to evaluate (line 13 in Algorithm \ref{alg:specialist}).

\paragraph{Specialist Method in Practice} The main constraint in development of a Specialist metric is the availability of ratings to use as ICL examples for the given test set. However, note that it is much cheaper and more efficient to collect a set of (human) ratings from a few translation systems for a single test set as a one-off investment, than to repeatedly depend on human annotators for evaluation of new translation models (e.g., throughout the model development process). Performance of this Specialist metric as a function of the number of ratings will be explored in \S\ref{ablation:icl_scaling}, where we show that ratings from only 3 translation systems are sufficient to exceed the state-of-the-art.
\section{Experimental Setup}
\label{sec:setup}

\subsection{Datasets}
\label{setup:datasets}

The Specialist method (described in \S\ref{sec:method}) depends on having access to a test set with multiple ratings (of different translations) for each input. Such ratings have already been collected as part of the Conference on Machine Translation (WMT) Metrics Shared Tasks in 2023 \citep{freitag2023results} and 2024 \citep{freitag2024llms}. We will refer to these datasets as WMT'23 and WMT'24, respectively.

We use the MQM ratings for English-German (en$\rightarrow$de) and Chinese-English (zh$\rightarrow$en) from the WMT'23 dataset, and the MQM ratings for English-German (en$\rightarrow$de), English-Spanish (en$\rightarrow$es), and Japanese-Chinese (ja$\rightarrow$zh) from the WMT'24 dataset. For all datasets, we exclude the human-generated references, so that our metrics are reference-free (i.e., QE). See Table \ref{tab:num_systems_wmt} in Appendix \ref{appendix:a} for the number of translation systems per language pair, for WMT'23 and WMT'24. Except for the en$\rightarrow$es WMT'24 dataset, all ratings were collected in a \textit{pseudo-SxS} fashion \citep{riley2024finding}, which means that a fixed rater was assigned to rate all translations of a given input. For the en$\rightarrow$es WMT'24 test set, on the other hand, this constraint was not enforced (i.e., raters were not assigned so as to ensure that translations of the same input were rated by the same rater).

\subsubsection{Additional Rounds of Ratings}
\label{setup:mqm-rounds}
In order to better understand how inter-rater variability affects the performance of the Specialist metric, we take advantage of additional rounds of MQM ratings. In \S\ref{sec:raters}, we use these supplemental ratings to show that that the Specialist metric specializes not only to the test set but also to the rater and, hence, it is important that ratings be collected in a pseudo-SxS manner. Note that \citet{riley2024finding} also found that collecting pseudo-SxS ratings is crucial for stability (i.e., replicability) of human evaluations.
\begin{itemize}
    \item \textbf{WMT'23 Round2 and Round3}: Two additional rounds of WMT'23 MQM ratings, rated by the same set of raters as in the first round, but with individual translations being assigned to strictly different raters in each round. As with the first round, the second two rounds of ratings were also collected in a pseudo-SxS fashion.
    \item \textbf{WMT'23 Multi-Rater Subset}: An extension to the (Round1) WMT'23 zh$\rightarrow$en MQM ratings, whereby 10\% of the test set (18 source segments $\times$ 15 systems = 270 examples) was rated by all 8 raters.
\end{itemize}

\subsection{Models}
\label{sec:models}
We use the Gemini 1.5 Pro model \citep{geminiteam2024geminifamilyhighlycapable} as the prompted LLM evaluator for all experiments (unless otherwise indicated; see \S\ref{ablation:gpt}). During creation of the Specialist ICL examples, we exclude the human reference, so all Specialist models (as well as baselines) are QE (i.e., reference-free) metrics.

The system and user instructions used for prompting AutoMQM are shown in Figure \ref{fig:automqm_prompt} in Appendix \ref{appendix:b}, and were adapted from the GEMBA instructions \citep{kocmi2023gemba}. As indicated in the prompt, the output is expected to be provided in JSON format, and all ICL examples are also provided in the expected output format, with each error having \texttt{span}, \texttt{severity}, and \texttt{category} fields. See Table \ref{tab:automqm_output} in Appendix \ref{appendix:b} for an example AutoMQM output.

\paragraph{Primary baselines}
We compare our proposed Specialist AutoMQM metric against the following baselines:
\begin{itemize}
    \item External baselines:
    \begin{itemize}
        \item XCOMET \citep{guerreiro2023xcomet}: Current state-of-the-art automatic metric for \textit{span-based} machine translation evaluation.
        \item GEMBA-MQM \citep{kocmi2023gemba}: Closest precedent to our proposed metric, which also prompts an LLM (GPT-4) for the task of MQM prediction. GEMBA uses a fixed set of 3 (English-German, English-Czech, and Chinese-English) ICL examples (unlike our Specialist metric, for which every test example is accompanied by a unique set of ICL examples).
        \item MetricX \citep{juraska2024metricx}: Current state-of-the-art automatic metric for machine translation evaluation and winner of the WMT'24 Metrics Shared Task \citep{freitag2024llms}. This model is only used as a baseline for the Specialist Scorer experiments in \S\ref{ablation:scorer}.
    \end{itemize}
    \item ``Shuffled sources'': The same global set of ICL examples (per test set) as the Specialist model is used, but these ICL examples are shuffled across test examples.
    \item ``Fixed, different source'': The same global set of ICL examples (per test set) as the Specialist model is used, but these ICL examples are permuted (relative to the Specialist setup) so that, for a given test example, all of its ICL examples come from a fixed source, which is strictly different than that of the test example, but has the same rater. See Figure \ref{fig:specialist_diagram} for an illustration of this setup versus the Specialist setup.

For both the ``Shuffled sources'' and ``Fixed, different source'' baselines, the following constraint is enforced: The ICL examples for a given test translation cannot include any translations, whether from the same or different source, produced by the same system as that which produced the test translation. Moreover, both of these baselines use the same number of ICL examples per test example as the Specialist model.
\end{itemize}

\setlength{\tabcolsep}{0.75em}
\begin{table*}[ht]
\centering
\begin{tabular}{l|ccc|cc}
\toprule
Character-level F1 & Same source as test & Fixed source & Same rater & WMT'23 & WMT'24 \\
\midrule
\multicolumn{1}{@{}l}{\raisebox{1.2ex}{\scriptsize \textbf{Baselines}}}\\[-6pt]
1a) XCOMET-XXL-QE & --- & --- & --- & 33.50\phantom{*} & 16.23\phantom{*} \\
1b) GEMBA-MQM-QE & \xmark & \xmark & \xmark & 31.99\phantom{*} & --- \\
1c) Shuffled sources & \xmark & \xmark & \xmark & 34.65\phantom{*} & 24.46\phantom{*} \\
1d) Fixed, different source & \xmark & \cmark & \cmark & 27.06\phantom{*} & 19.96\phantom{*}\\
\midrule
2a) Specialist & \cmark & \cmark & \cmark & \textbf{51.59}$^*$ & \textbf{35.59}$^*$ \\
\bottomrule
\end{tabular}
\vspace{-0.5em}
\caption{Character-level F1 on the WMT'23 and WMT'24 test sets for Specialist and Baseline metrics, averaged over all language pairs per test set. See Table \ref{tab:main_table_wmt23} in Appendix \ref{appendix:a} for F1, precision, and recall, broken out by language pair. See \S\ref{sec:models} for a description of all of the Baseline and Specialist systems. For WMT'23, the ``Shuffled sources'' and ``Fixed, different source'' results are computed as the average over 10 runs with different random seeds. See Table \ref{tab:baselines_avg_variance} in Appendix \ref{appendix:a} for the variance across runs. Asterisks ($*$) indicate scores which are statistically significantly better than XCOMET (row 1a), according to a paired permutation test.
\label{tab:main_table_wmt23_wmt24}
}
\end{table*}

\subsection{Creation of Specialist ICL examples}
The Specialist metric (as described in \S\ref{sec:method}) evaluates the quality of a single translation system $M^{*}$ given known ratings for a set of $N$ other translation systems $\left\{M_j\right\}_{j=1:N}$. Here, for each test set (WMT'23 and WMT'24, across all language pairs), we have access to ground-truth ratings for all system outputs. In order to meta-evaluate the Specialist AutoMQM metric, we first collect predictions from this metric for each system, via hold-one-out prompting; that is, for whichever system we are evaluating, we exclude that system's ratings from the ICL examples and prompt with the ratings from the remaining systems. Thus, the number of ICL examples per test example is equal to the total number of systems $- 1$ (excluding the system which generated the test example). Then, we gather the Specialist AutoMQM predictions across all systems to perform meta-evaluation of this metric over the entire test set of interest. (Note that we also include a performance breakdown by system in \S\ref{ablation:systems}.) See Tables \ref{tab:error_stats_23} (WMT'23) and \ref{tab:error_stats_24} (WMT'24) in Appendix \ref{appendix:a} for the average number of ICL examples and average number of total errors across ICL examples per test example.

\subsection{Meta-evaluation}

To meta-evaluate the quality of Specialist AutoMQM, we compute the \textit{character-level precision, recall, and F1} span tagging evaluation metrics (used by the WMT'23 QE Shared Task; \citet{blain2023findings}). Given gold and predicted ratings, these metrics represent the precision, recall, and F1 of predicting whether a character in the hypothesis translation is included in an error span or not. Partial credit of 0.5 is given if the predicted rating marks a character as an error but predicts the incorrect severity.

To meta-evaluate the Specialist Scorer (which is not a span-based metric) in \S\ref{ablation:scorer}, we report segment-level pairwise accuracy with tie calibration (which we refer to as \textit{Acc23}; \citet{deutsch2023ties}), which is the segment-level meta-evaluation metric used in the WMT'23 and WMT'24 Metric Shared Tasks. Acc23 rewards metrics for correctly ranking translations as well as correctly predicting ties, in combination with a tie calibration procedure that automatically introduces ties into metric scores so that the meta-evaluation is fairer.
\section{Results and Discussion}
\label{sec:results}
The main results are shown in Table \ref{tab:main_table_wmt23_wmt24}. (See Tables \ref{tab:main_table_wmt23} and \ref{tab:main_table_wmt24} in the Appendix \ref{appendix:a} for a breakdown of the results by language pair for WMT'23 and WMT'24, respectively, with F1, precision, and recall metrics.) First note that the ``Shuffled sources'' baseline (row 1c) already performs on par with the existing state-of-the-art AutoMQM models (XCOMET-XXL-QE in row 1a and GEMBA-MQM-QE in row 1b) on the WMT'23 test set, and and outperforms XCOMET on the WMT'24 test set.
Also note that the ``Fixed, different source'' baseline (row 1d) underperforms ``Shuffled sources'', due almost entirely to a drop in recall. This may be because ICL examples in the ``Shuffled sources'' baseline sometimes include examples with the same source as the test example, while for the ``Fixed, different source'' baseline, we enforce that the source must be different than that of the test example.

In contrast, the ``Specialist'' setting (row 2a) dramatically outperforms all of the baselines, with a 54\% improvement in F1 score (averaged over both language pairs) relative to XCOMET (row 1a), the current state-of-the-art. The improvement is even more dramatic for WMT'24, with the ``Specialist'' achieving a relative 119\% improvement in F1 score relative to XCOMET (rows 1a vs 2a in Table \ref{tab:main_table_wmt24}). Recall that the difference between the ``Specialist'' setting and the ``Fixed, different source'' setting is that, in the former setting, error annotations from translations of the \textit{same source} as the test example are provided as demonstrations. Thus, same-source demonstrations are crucial to the success of our method, and its success cannot be attributed only to (i) providing demonstrations of errors from different translations of some \textit{fixed source}, or (ii) providing ICL examples from the \textit{same rater} as the test translation rating ground truth.

In the following sections, we present a series of ablations and analyses designed to understand the representations learned by our Specialist metrics, and to verify the generalizability and robustness of our Specialist method for designing automatic metrics across different numbers and distributions of ICL examples, different LLM backbones, and different evaluation tasks.

\begin{figure*}[t]
\centering
\subfigure[WMT'23 en-de]{\includegraphics[scale=0.17]{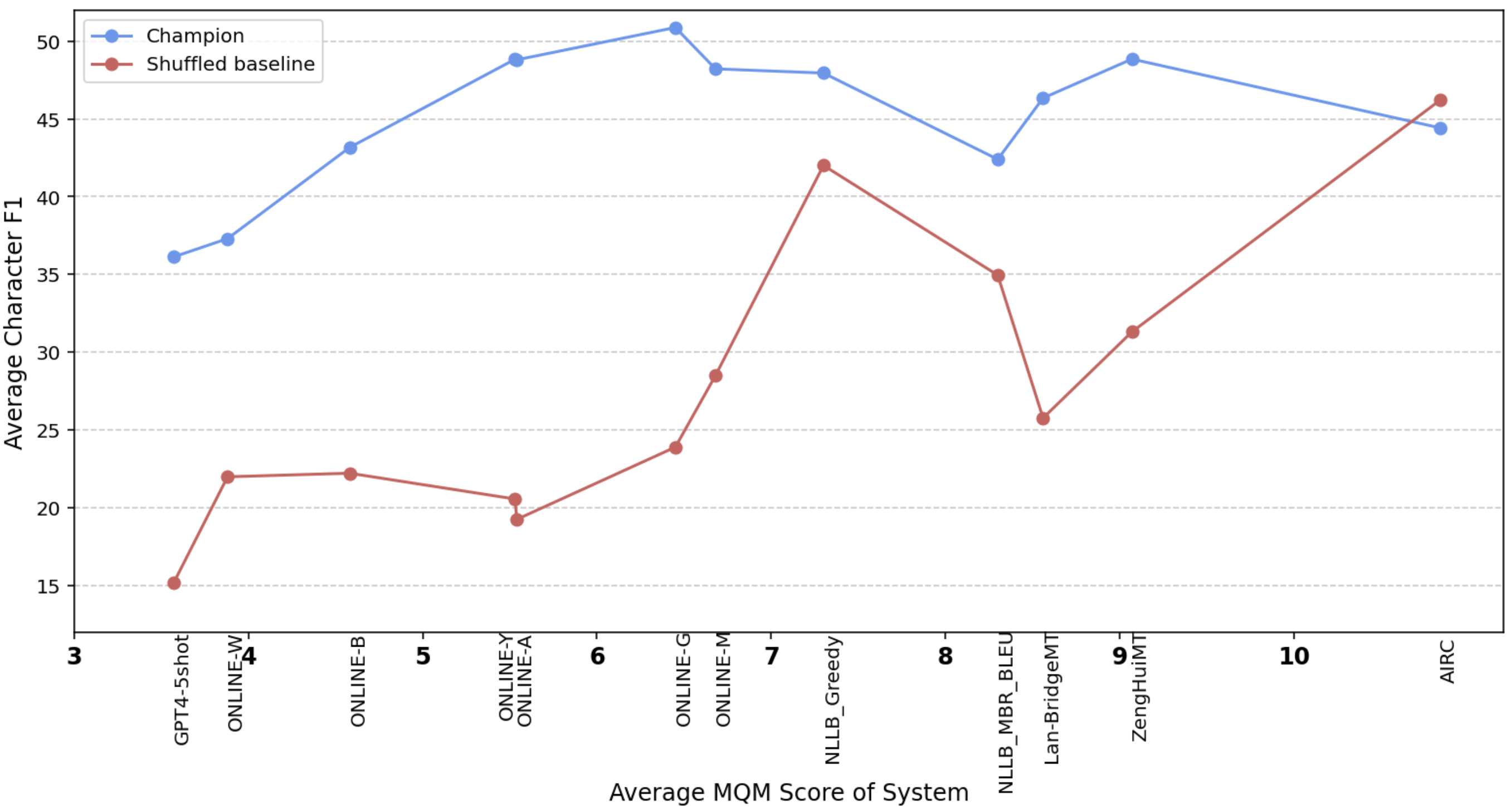}}\quad
\subfigure[WMT'23 zh-en]{\includegraphics[scale=0.17]{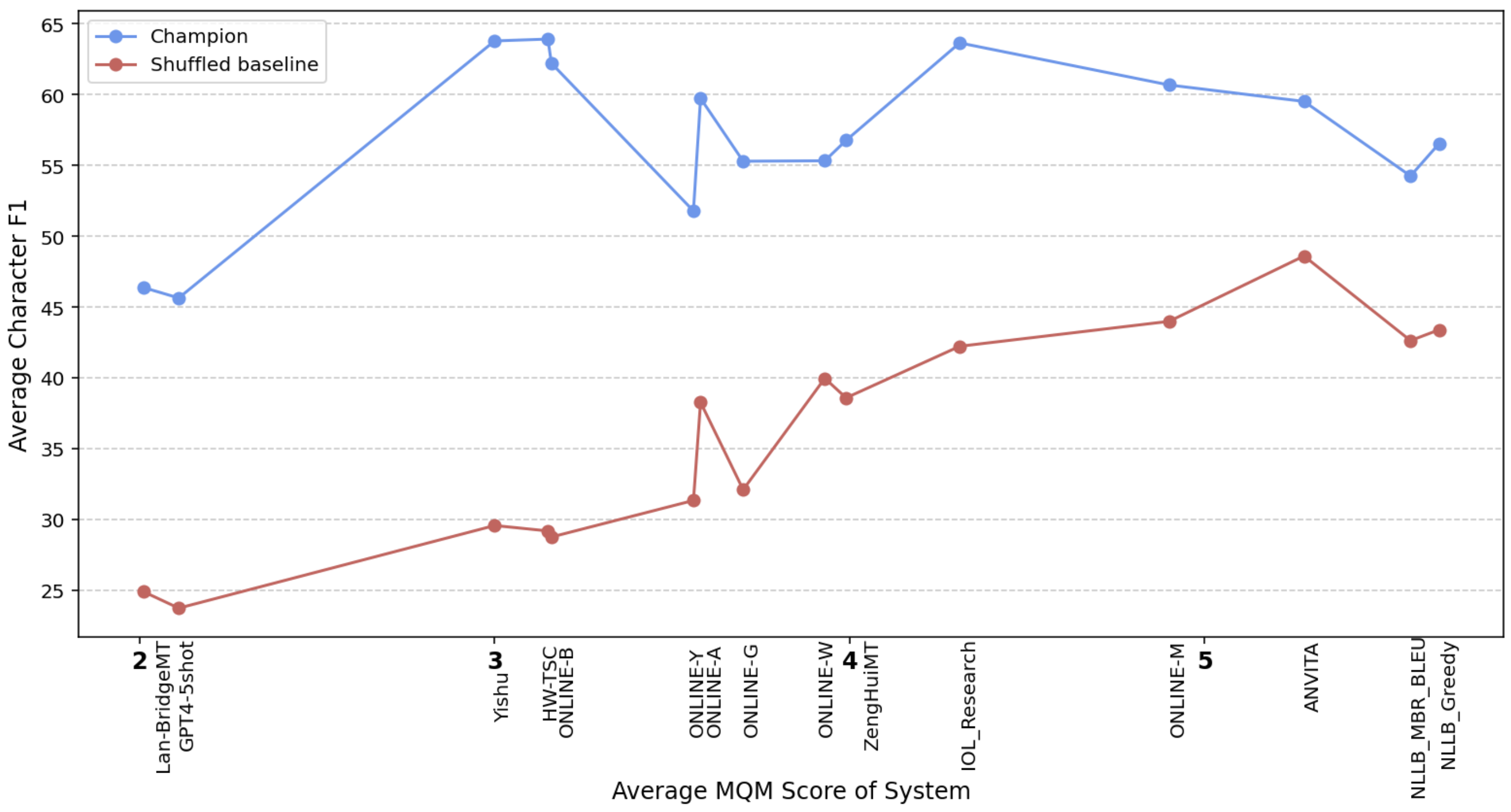}}
\caption{Specialized AutoMQM performance per translation system. The Champion (row 2a) and Shuffled baseline (row 1c) models from Table \ref{tab:main_table_wmt23} are compared. The average MQM score of each system, along with its name, is shown in the x-axis. The average character-level F1 of the AutoMQM model when evaluated on this system only is shown on the y-axis.}
\label{fig:system_level_meta_eval}
\end{figure*}

\subsection{Is Specialist AutoMQM Robust to the Choice of LLM?}
\label{ablation:gpt}
The results reported in Table \ref{tab:main_table_wmt23_wmt24} used the Gemini 1.5 Pro LLM, and showed huge gains of Specialist AutoMQM over the shuffled baseline. To investigate whether gains from this method generalize to other LLMs, we also performed the same comparison using the GPT-4o \citep{achiam2023gpt} and Claude 3.5 Sonnet \citep{bai2022constitutional} LLMs. As with Gemini 1.5 Pro, Specialist AutoMQM also substantially outperforms Shuffled AutoMQM when prompting these other LLMs (Table \ref{tab:gpt_vs_gemini}). For example, using GPT-4o, the character-level F1 score increases from 33.6 to 48.3 for WMT'23 en$\rightarrow$de, and from 38.4 to 56.8 for WMT'23 zh$\rightarrow$en. Moreover, note that Specialist AutoMQM with the GPT-4o backbone outperforms GEMBA-MQM-QE (which is also a prompted GPT-4 model, albeit an earlier version) by an even larger margin. This supports the effectiveness of our approach over baselines using external (different-source) ICL examples. In the remaining experiments, we continue to use the Gemini 1.5 Pro LLM.
\setlength{\tabcolsep}{0.5em}
\begin{table}[ht]
\centering
\begin{tabular}{lc|c}
\toprule
Character-level F1 & en$\rightarrow$de & zh$\rightarrow$en \\
\midrule
\multicolumn{1}{@{}l}{\raisebox{1.2ex}{\scriptsize \textbf{Baselines}}}\\[-6pt]
XCOMET-XXL-QE & 32.71 & 34.29 \\
GEMBA-MQM-QE & 29.80 & 34.17 \\
\midrule[0.05pt]
\multicolumn{1}{@{}l}{\raisebox{1.2ex}{\scriptsize \textbf{Gemini-AutoMQM}}}\\[-6pt]
Shuffled sources & 31.12 & 37.62 \\
Specialist AutoMQM & 45.71 & \textbf{57.47} \\
\midrule[0.05pt]
\multicolumn{1}{@{}l}{\raisebox{1.2ex}{\scriptsize \textbf{GPT-4o-AutoMQM}}}\\[-6pt]
Shuffled sources & 33.58 & 38.38\\
Specialist AutoMQM & 48.32 & 56.78 \\
\midrule[0.05pt]
\multicolumn{1}{@{}l}{\raisebox{1.2ex}{\scriptsize \textbf{Claude-3.5-Sonnet-AutoMQM}}}\\[-6pt]
Shuffled sources & 35.02 & 40.86 \\
Specialist AutoMQM & \textbf{48.49} & 56.13 \\
\bottomrule
\end{tabular}
\caption{Comparison of Specialist AutoMQM vs the shuffled baseline (WMT'23 test set) for three different LLMs: Gemini 1.5 Pro, GPT-4o, and Claude-3.5-Sonnet. For all three LLMs, Specialist AutoMQM substantially outperforms the shuffled baseline.
\label{tab:gpt_vs_gemini}
}
\end{table}

\subsection{How Does Specialist AutoMQM Performance Vary Across Translation Systems?}
\label{ablation:systems}

In practice, the Specialist AutoMQM metric would likely be used to evaluate the quality of a single translation system, or to compare a pair of systems, given historical ratings from other systems. The WMT'23 and WMT'24 MQM test sets contain rated translations from at least a dozen translation systems per language pair (Table \ref{tab:num_systems_wmt}). These systems are of varying quality, and aggregate meta-evaluation of AutoMQM (Table \ref{tab:main_table_wmt23_wmt24}) could hide per-system differences in metric performance. Here, we compare performance of Specialist AutoMQM against the ``Shuffled sources'' baseline on a per-system basis for WMT'23. As shown in Figure \ref{fig:system_level_meta_eval}, Specialist AutoMQM outperforms the shuffled baseline for every zh$\rightarrow$en system, and for every en$\rightarrow$de system except the lowest-quality one. Thus, Specialist AutoMQM outperformance is consistent for translation systems across the quality spectrum, and cannot be explained by outperformance only for a certain translation quality tier. Note that both the Specialist and shuffled baseline AutoMQM models tend to perform worse on the highest-quality systems (e.g. GPT4-5shot), likely due to limitations in the underlying translation capabilities of the backbone language model used for AutoMQM.

\begin{figure*}[t]
\centering
\subfigure[WMT'23 en-de]{\includegraphics[scale=0.25]{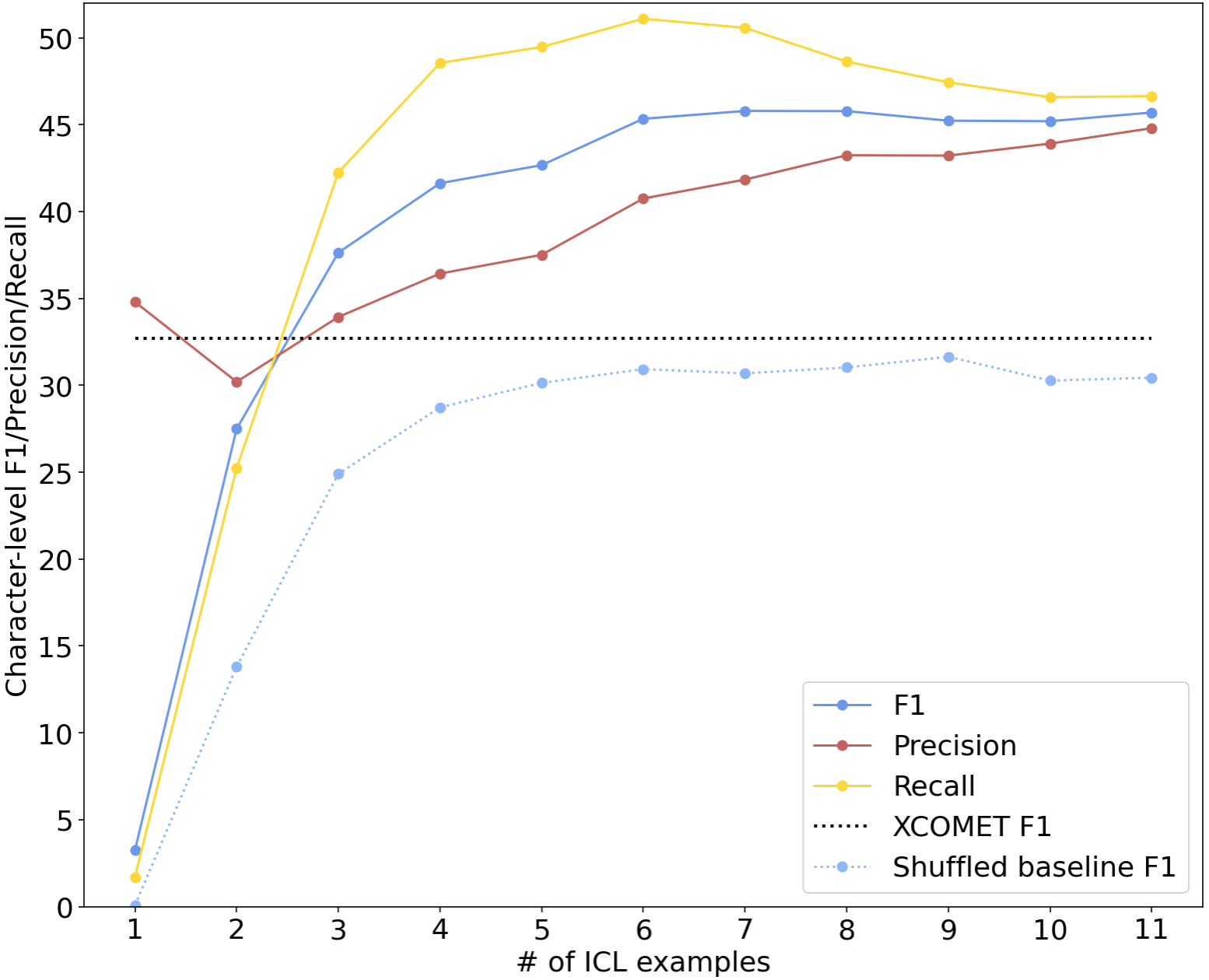}}\quad
\subfigure[WMT'23 zh-en]{\includegraphics[scale=0.25]{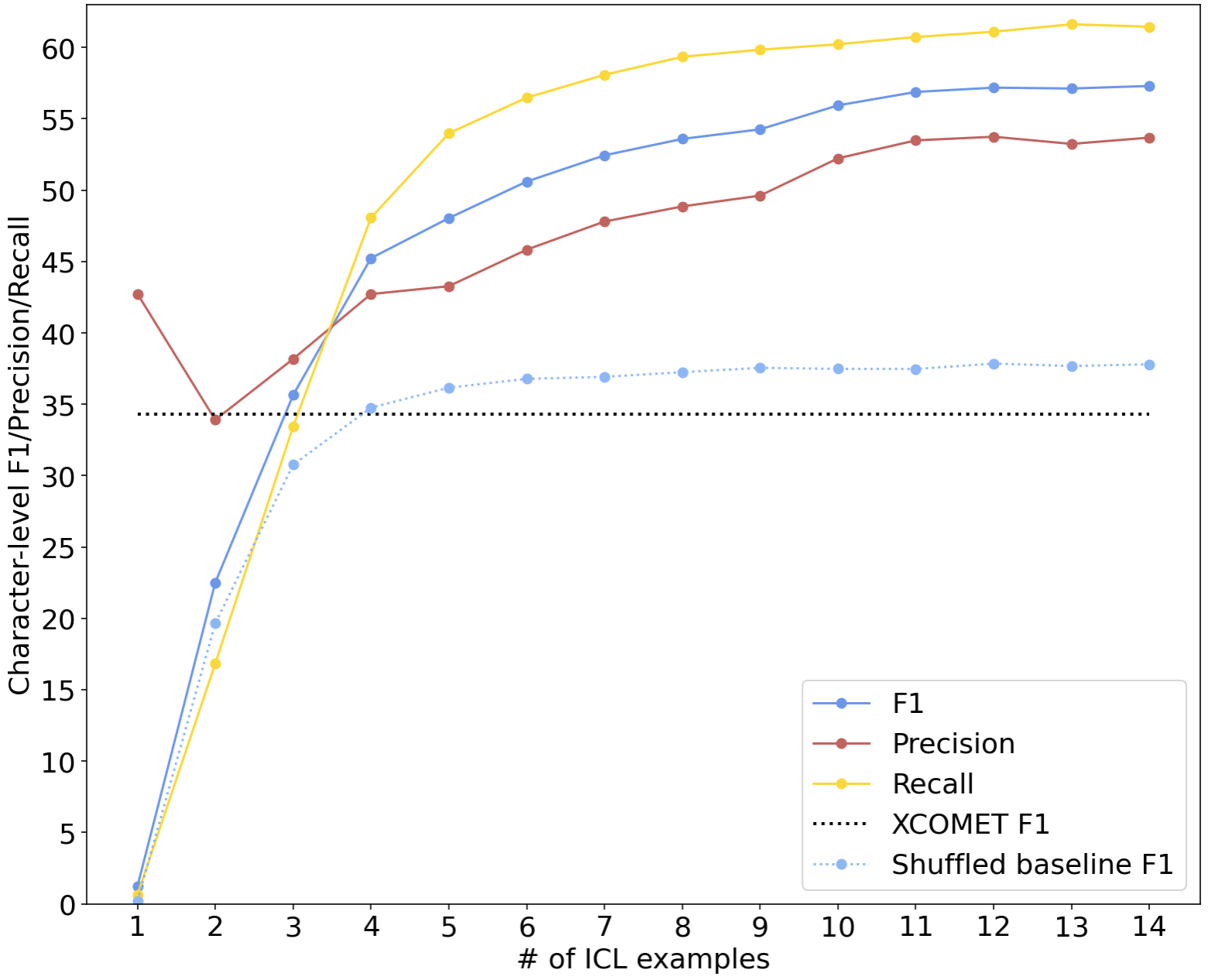}}
\caption{Specialist AutoMQM performance (``Champion + Filter'' setting; Table \ref{tab:main_table_wmt23}, row 2b) as a function of number of ICL examples used. For comparison, XCOMET performance, as well as ICL example scaling for the ``Shuffled sources'' baseline (Table \ref{tab:main_table_wmt23}, row 1c), are also shown.}
\label{fig:icl_examples_scaling}
\end{figure*}

\subsection{How Does Specialist AutoMQM Performance Scale as a Function of Number of ICL Examples?}
\label{ablation:icl_scaling}

The most expensive and time-consuming step in the Specialist AutoMQM process is collecting ratings to use as demonstrations. Thus, it is useful to understand the marginal improvements in performance that can be expected as a result of collecting additional ratings. In this ablation, we randomly select subsets of Specialist AutoMQM ICL examples in the range [1, num\_systems - 1], inclusive. (See Table \ref{tab:num_systems_wmt} for the total number of systems for each language pair in the WMT'23 and WMT'24 test sets.) While scaling ICL examples, we incrementally add a single new example to the existing set (for each test example), so that every set of ICL examples of a given size $n$ is a superset of the ICL examples for all sizes less than $n$. Also note that when the number of ICL examples reaches its maximum (of num\_systems - 1), this  corresponds to the results reported in row 2a of Table \ref{tab:main_table_wmt23_wmt24}.

As shown in Figure \ref{fig:icl_examples_scaling}, increasing the number of ICL examples improves character-level F1 monotonically up to 7 ICL examples for en$\rightarrow$de, and up to 12 ICL examples for zh$\rightarrow$en. For reference, we show the XCOMET-XXL-QE performance as a baseline in Figure \ref{fig:icl_examples_scaling}, and also show results from the same scaling experiment using the ``Shuffled sources'' setting (Table \ref{tab:main_table_wmt23_wmt24}, row 1c). We see that only 3 ICL examples are needed for Specialist AutoMQM to outperform XCOMET, for both en$\rightarrow$de and zh$\rightarrow$en. We also see that Specialist AutoMQM outperforms the shuffled baseline at every ICL example set size, and that the benefit from increasing number of ICL examples plateaus more quickly for the shuffled baseline than for Specialist AutoMQM.

Also see Table \ref{tab:additional_icl_examples} in Appendix \ref{appendix:a}, which shows no improvement from augmenting the (num\_systems - 1) same-source ICL examples with other (different-source) ICL examples from the same test set. This result, together with the performance saturation observed in Figure \ref{fig:icl_examples_scaling}, suggests that there is limited headroom to improve AutoMQM by further filling up the LLM's long context window (either with same-source or different-source examples).

\subsection{Is Specialist AutoMQM Simply Copying Errors From ICL Examples?}
\label{ablation:copy}

\setlength{\tabcolsep}{0.45em}
\begin{table*}
\centering
\begin{tabularx}{\linewidth}{Lcc|cc}
\toprule
& \small{3 ICL Examples} & \small{11 ICL Examples} & \small{Shuffled} & \small{11 ICL Examples} \\
\midrule
1) \small{Total predicted error count} & 14,539 & 14,481 & 10,298 & 14,481 \\
2) \small{Disjoint error count} & 9,185 & 9,127 & 7,913 & 12,096 \\
3) \small{Disjoint error count with exact match to ICL example errors} & 4,142 & 2,327 & 6,374 & 494 \\
\bottomrule
\end{tabularx}
\caption{Pairwise comparison of predicted errors copied from ICL examples, for different AutoMQM systems. The left two columns show a comparison of the Specialist AutoMQM (Table \ref{tab:main_table_wmt23_wmt24}, row 2a) prompted with 3 vs 11 ICL examples, and the right two columns show a comparison of the latter against the shuffled baseline (Table \ref{tab:main_table_wmt23_wmt24}, row 1c). Row 1 shows the total predicted error count over the full WMT'23 en$\rightarrow$de test set, row 2 shows the number of errors predicted by the given system which were not predicted by the other system being compared, and row 3 shows the subset of these errors which are exact matches to errors from all 11 (same-source) ICL examples.}
\label{tab:icl_copy_count}
\end{table*}
\subsubsection{Learning When to Abstain}
In view of the results from \S\ref{ablation:icl_scaling}, i.e., that increasing the number of ICL examples improves performance of Specialist AutoMQM, this raises the question of whether the performance improvements are simply due to the model copying errors that it is shown in the ICL examples. First, observe that the performance improvements shown in Figure \ref{fig:icl_examples_scaling} (\S\ref{ablation:icl_scaling}) from increasing the number of ICL examples are eventually due to improvements in precision, while recall flattens out (for zh$\rightarrow$en) or even starts to decrease (for en$\rightarrow$de). This suggests that the model becomes more conservative in its predictions when provided with more ICL examples (rather than increasing the rate at which it copies from ICL examples).

As shown in Table \ref{tab:icl_copy_count} (left two columns), when comparing Specialist AutoMQM prompted with 3 versus 11 ICL examples, once errors predicted by both AutoMQM systems are removed, the AutoMQM system prompted with 11 ICL examples predicts \textit{fewer} errors which are direct copies of spans from all 11 ICL examples than the AutoMQM system prompted with 3 ICL examples. The comparison is even more pronounced when comparing the ``Shuffled Sources'' baseline against Specialist AutoMQM with 11 ICL examples (right two columns of Table \ref{tab:icl_copy_count}). After removing errors predicted by both systems, the shuffled baseline predicts 6,374 errors which are direct copies of spans from the 11 ICL examples, while Specialist AutoMQM only predicts 494 such errors (even though Specialist AutoMQM predicts more total errors than the shuffled baseline). Thus, the AutoMQM system prompted with 11 ICL examples is \textit{abstaining} from predicting certain errors that it is shown via these demonstrations, while the shuffled baseline, which has not been shown these errors, is predicting them more liberally. This suggests that ICL examples not only teach the Specialist AutoMQM model which errors to predict, but also which parts of the translation are error-free. 

\subsubsection{Parrot Model Baseline}
To exactly quantify how much of Specialist AutoMQM's performance can be attributed to error copying from ICL examples, we construct an artificial baseline model, which we call the ``Parrot''. This model has access to the same ICL examples as Specialist AutoMQM, and makes predictions as follows: For every error present in ICL examples for which there is a matching span in the test translation, predict this as an error.

The results comparing the Parrot model with Specialist AutoMQM are shown in Table \ref{tab:copy_ablation}. Note that character-level F1 improves dramatically from 27.6 for the Parrot model to 45.7 for Specialist AutoMQM for en$\rightarrow$de, and from 36.5 to 57.5 for zh$\rightarrow$en. Thus, the performance of Specialist AutoMQM cannot be solely explained by naive copying behavior.

The gap in recall between the Parrot model and Specialist AutoMQM (29.9 vs 46.4 for en$\rightarrow$de, and 38.3 vs 61.4 for zh$\rightarrow$en) quantifies the extent to which Specialist AutoMQM correctly predicts errors not present in ICL examples. Observe that, even when copying the maximum possible number of errors from ICL examples (whose span is also present in the test translation), there is still a large gap in recall (i.e., many correct errors that the Parrot fails to predict). See examples in Table \ref{tab:new_error_pred_examples} (Appendix \ref{appendix:a}) of where AutoMQM correctly identifies errors not present in ICL examples.

The gap in precision between the Parrot model and Specialist AutoMQM (25.6 vs 45.0 for en$\rightarrow$de and 34.9 vs 54.1 for zh$\rightarrow$en) quantifies the extent to which Specialist AutoMQM correctly abstains from predicting errors present in ICL examples. There are many cases where an error span present in ICL examples does not correspond to a (ground-truth) error in the test translation, and Specialist AutoMQM is able to identify many of these. See examples of where AutoMQM correctly abstains from predicting errors present in ICL examples in Table \ref{tab:copy_abs_examples} (Appendix \ref{appendix:a}).

\setlength{\tabcolsep}{0.5em}
\begin{table}[ht]
\centering
\begin{tabular}{lc|c}
\toprule
Character-level F1 & en$\rightarrow$de & zh$\rightarrow$en \\
\midrule
Shuffled sources & 31.12& 37.62 \\
Parrot & 27.59 & 36.52 \\
Specialist AutoMQM & 45.71 & 57.47 \\
\bottomrule
\end{tabular}
\vspace{-0.5em}
\caption{Comparison of ``Parrot model'' performance against that of Specialist AutoMQM (WMT'23 test set). The Parrot has access to the same ICL examples as Specialist AutoMQM, and makes predictions as follows: For every error present in ICL examples for which there is a matching span in the test translation, predict this as an error.
\label{tab:copy_ablation}
}
\end{table}

\begin{table*}[ht]
\centering
\begin{tabular}{lcccccc}
\toprule
& \multicolumn{3}{c}{en$\rightarrow$de} & \multicolumn{3}{c}{zh$\rightarrow$en} \\
\cmidrule{2-4} \cmidrule{5-7}
& F1 & Precision & Recall & F1 & Precision & Recall \\
\midrule
\multicolumn{1}{@{}l}{\raisebox{1.2ex}{\scriptsize \textbf{Human agreement}}}\\[-6pt]
1a) Round2 & 34.91 & 38.16 & 32.17 & 38.68 & 39.00 & 38.36 \\
1b) Round3 & 38.46 & 40.26 & 36.82 & 39.16 & 40.06 & 38.29 \\
\midrule
\multicolumn{1}{@{}l}{\raisebox{1.2ex}{\scriptsize \textbf{Specialist}}}\\[-6pt]
2a) Round1 ICL & 45.71 & 45.04 & 46.40 & 57.47 & 54.05 & 61.36 \\
2b) Round2 ICL & 30.80 & 30.87 & 30.74 & 38.83 & 36.65 & 41.30 \\
2c) Round2 $\vert$ Round3 ICL & 30.83 & 26.65 & 36.56 & 38.48 & 29.91 & 53.93 \\
\bottomrule
\end{tabular}
\vspace{-0.5em}
\caption{Specialist AutoMQM performance when prompting and evaluating using different raters (WMT'23 test set). ``Round 2 $\vert$ Round3'' indicates that ratings from these rounds were merged. Results are reported using the Round1 test set (which is the same test set used in all other tables, and in official WMT'23 results.)
\label{tab:rater_ablations}
}
\end{table*}
\subsection{Is Specialist AutoMQM Specialized Only to a Test Set, or Also to a Rater?}
\label{sec:raters}

By construction, Specialist AutoMQM is a specialist for a given test set. Since the WMT'23 and WMT'24 test sets are all constructed in a pseudo-SxS fashion (with the exception of WMT'24 en$\rightarrow$es), Specialist AutoMQM is also prompted with ICL examples rated by the same rater as the test translation. Here, we seek to understand whether Specialist AutoMQM specializes only to the test set, or also to the rater. To answer this question, we take advantage of the additional rounds of WMT'23 MQM ratings, as described in \S\ref{setup:mqm-rounds}.

\paragraph{Prompting with a Different Rater} In the first set of experiments, we use the same-source Specialist AutoMQM set-up, but prompt with ICL examples taken from Round2 ratings. We always evaluate using the official Round1 ratings. As shown in Table \ref{tab:rater_ablations} (row 2b), performance drops to the level of the ``Shuffled sources'' baseline (Table \ref{tab:main_table_wmt23}, row 1c) when prompting with these Round2 ICL examples. Thus, Specialist AutoMQM specializes both to the test set and to the rater (on a per-example basis; recall that with the pseudo-SxS setup, raters can still vary across different inputs). This is not surprising, since there are large (and often competing) differences in behavior across raters, and AutoMQM performance is being measured according to the judgments from a single (Round1) rater per example. See rows 1a) and 2a) in Table \ref{tab:rater_ablations} for the inter-annotator agreement across rounds (Round2 vs Round1, and Round3 vs Round1, respectively). Note that the Round2 Specialist (row 2b) performs on par with the inter-annotator agreement for zh$\rightarrow$en, and the Round1 Specialist (row 2a) outperforms the inter-annotator agreement for both language pairs, likely because it is able to match specific rater behavior from the ICL examples.
\begin{figure*}[t]
\centering
\subfigure[AutoMQM Character F1]{\includegraphics[scale=0.17]{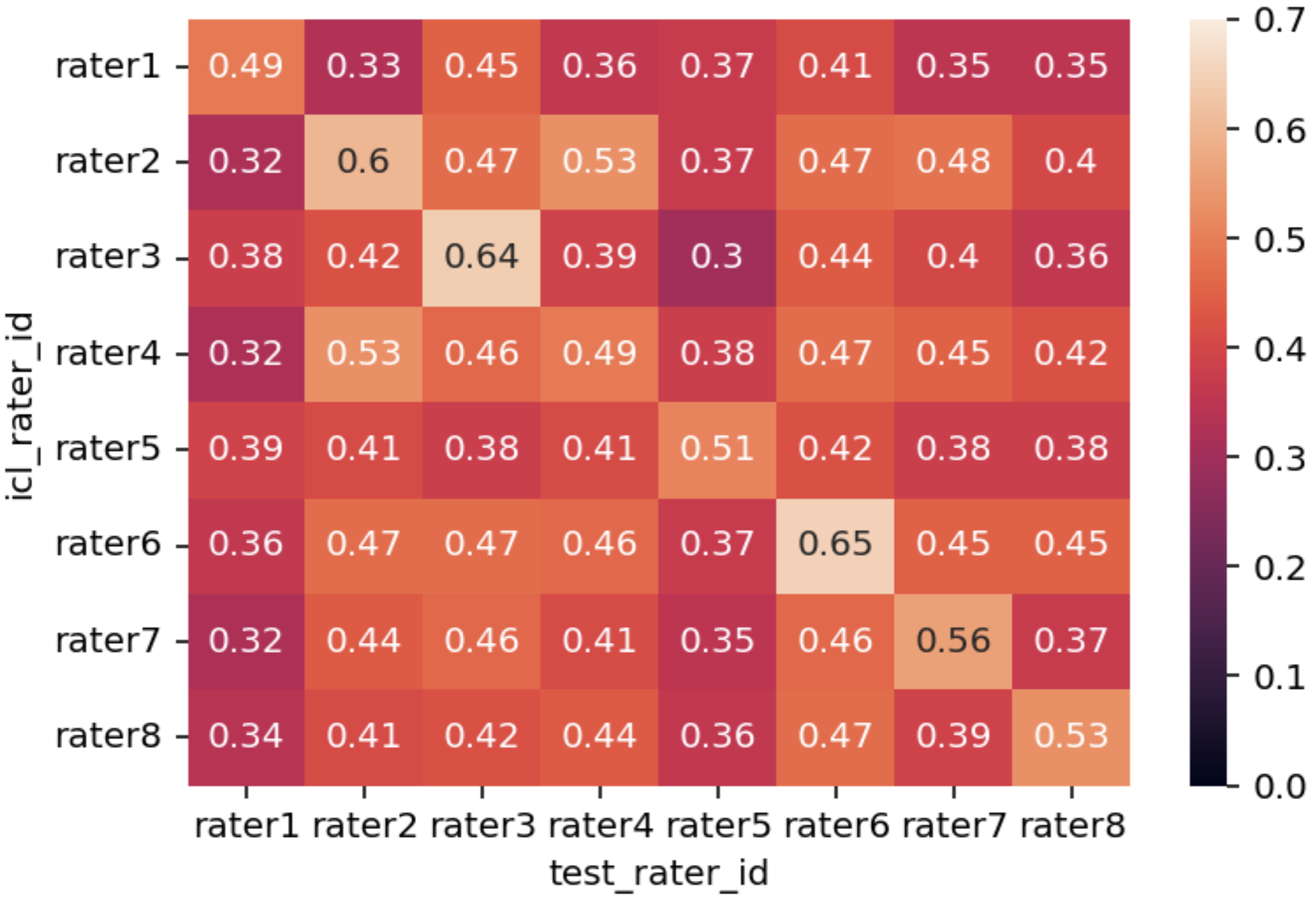}}
\subfigure[Inter-Annotator Character F1]{\includegraphics[scale=0.17]{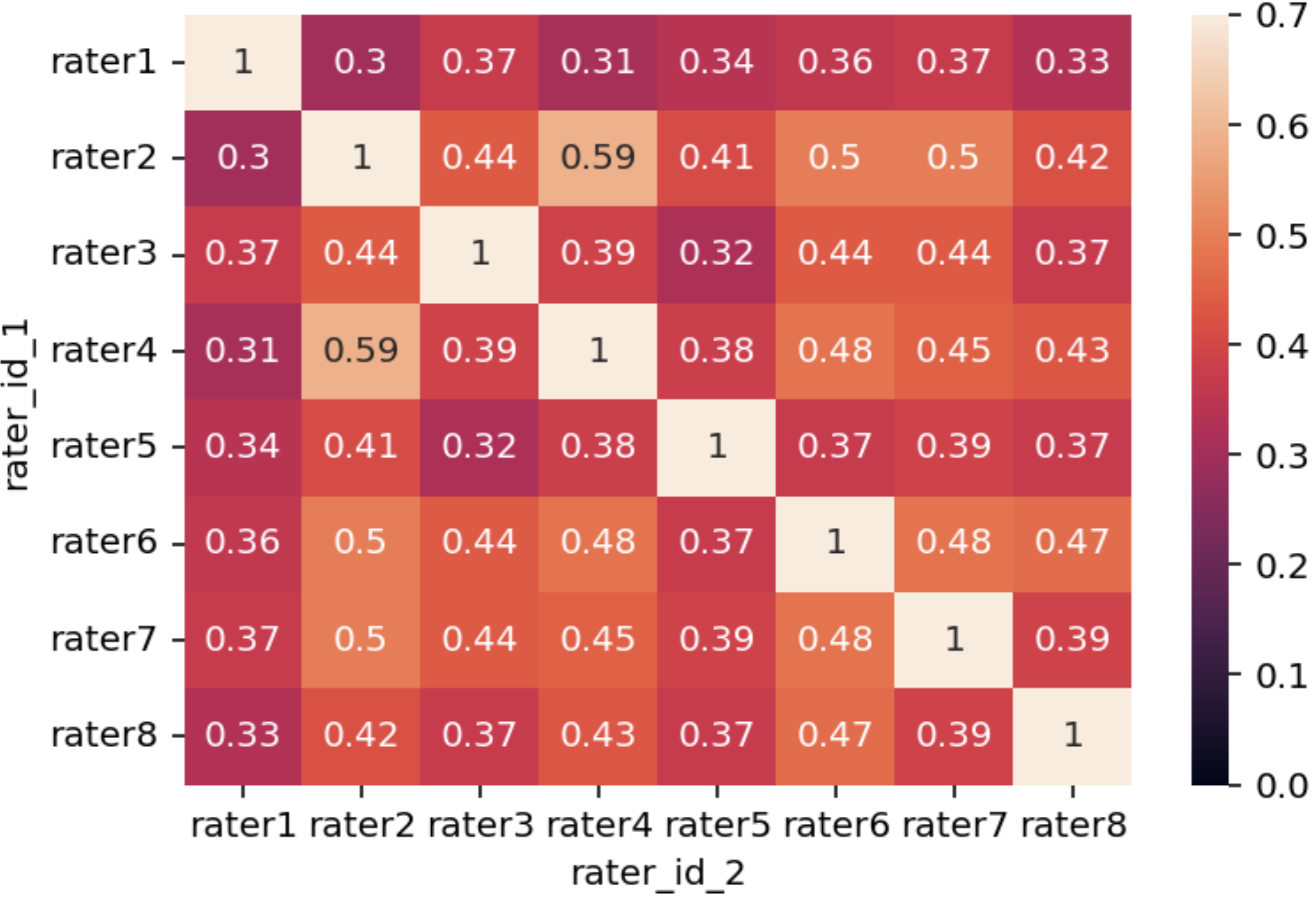}}\quad
\caption{Cross-rater performance of AutoMQM and human annotators, computed using the extension to the (round 1) WMT'23 zh$\rightarrow$en test set, whereby 10\% of the test set (18 source segments $\times$ 15 systems = 270 examples) was rated by all 8 raters. In Figure (a), Specialist AutoMQM is prompted with (same-source) ICL examples from the icl\_rater\_id rater (vertical axis), and evaluated using the ratings from the test\_rater\_id rater (horizontal axis). This figure shows the entire matrix of character-level F1 scores for every (icl\_rater\_id, test\_rater\_id) pair. In Figure (b), the matrix of character-level F1 scores between all pairs of human annotators is shown.
}
\label{fig:cross_rater_matrix}
\end{figure*}

\paragraph{Prompting with Merged Ratings} In a follow-up experiment, we merge ratings from Round2 and Round3, and provide these merged ratings as ICL examples (Table \ref{tab:rater_ablations}, row 2c). While character-level F1 is almost identical to when using Round2 ratings only as ICL examples (row 2b), we see that merging ratings across rounds improves recall (at the cost of lower precision). When comparing prompting just from Round2 ratings, versus Round2 and Round3 merged, recall increases from 30.7 to 36.6 for en$\rightarrow$de, and from 41.3 to 53.9 for zh$\rightarrow$en. The drop in precision when using merged ratings as ICL examples likely represents not an actual quality drop, but under-annotation of errors (low recall) by the Round1 raters (used as the test set). It remains an open question how to combine ratings from multiple raters to create a better test set, which would allow for measuring real improvements in Autorater quality that could not be captured by comparing against the ground-truth ratings from a single (imperfect) human rater per example.

\paragraph{ICL Rater $\times$ Test Set Rater Comparison} The aggregate results reported in Table \ref{tab:rater_ablations} could mask individual cross-rater dynamics since, for each round of ratings, the test set was split into chunks of equal size, and every chunk was rated by one of 10 en$\rightarrow$de raters and 8 zh$\rightarrow$en raters. While we do not have access to ratings from all raters for every system output across all test set examples, we do have access to the WMT'23 Multi-Rater Subset ratings (\S\ref{setup:mqm-rounds}), for which all 8 raters rated 10\% of the zh$\rightarrow$en dataset. We use this WMT'23 Multi-Rater Subset to understand cross-rater AutoMQM performance for all pairs of raters (where one rater is the ICL example annotator and the other rates the test set example), by computing the entire num\_raters $\times$ num\_raters matrix of F1 scores for every (ICL rater, test set rater) pair. These results are shown in Figure \ref{fig:cross_rater_matrix}(a). As expected, the F1 scores on the matrix diagonal are highest (though note that rater2 and rater4 also have high agreement). (Also see Table \ref{tab:cross_rater_f1} for the cross-rater performance using the full Round1 vs Round2 WMT'23 en$\rightarrow$de test sets, though note that for these rounds of ratings, we are missing entries for most of the $(\text{num\_raters})^2$ (ICL rater, test set rater) pairs.)

For comparison against AutoMQM, Figure \ref{fig:cross_rater_matrix}(b) shows the inter-annotator agreement (character-level F1) over the same Multi-Rater Subset of the zh$\rightarrow$en WMT'23 test set. Note that when prompting and evaluating with different raters, Specialist AutoMQM agrees with the raters about as much as the raters agree with each other.

\subsection{Does the Specialist Method Generalize to Other Automatic Evaluation Tasks?}
\label{ablation:scorer}

We have seen that the Specialist method for prompting LLMs-as-Judges achieves state-of-the-art performance for the task of MQM (fine-grained translation quality annotation). Here, we consider the task of scoring translation quality (without providing error annotations). (This task is also known in human evaluation parlance as ``direct assessment''.) In this task, we prompt the LLM to generate a float quality score on a scale from 0-100. (See Figure \ref{fig:da_prompt} in Appendix \ref{appendix:b} for the prompt used.)

As shown in Table \ref{tab:metricx_vs_automqm_23_24}, the task of score prediction also benefits substantially from the Specialist method (relative to the shuffled baseline). Here, we report segment-level accuracy, and also compare against MetricX-24 \citep{juraska2024metricx}, the state-of-the-art automatic score prediction metric for machine translation.

Note that Specialist AutoMQM outperforms the Specialist Scorer, perhaps because LLMs are better at natural language text generation versus generation of numbers. Thus, Specialist AutoMQM has a quality advantage over the Specialist Scorer, while also having the added benefit of interpretability. Also note that both Specialist models outperform MetricX-24 across all WMT'23 and WMT'24 language pairs except WMT'23 en$\rightarrow$de.
\setlength{\tabcolsep}{0.15em}
\begin{table}[ht]
\centering
\begin{tabular}{lcc|ccc}
\toprule
& \multicolumn{2}{c}{WMT'23} & \multicolumn{3}{c}{WMT'24} \\
\cmidrule{2-3} \cmidrule{4-6}
Segment-level Acc23 & en$\rightarrow$de & zh$\rightarrow$en & en$\rightarrow$de & en$\rightarrow$es & ja$\rightarrow$zh \\
\midrule
MetricX-24-QE & 59.44 & 54.48 & 52.45 & 68.48 & 52.69 \\
\midrule[0.05pt]
\multicolumn{1}{@{}l}{\raisebox{1.2ex}{\scriptsize \textbf{AutoMQM}}}\\[-6pt]
Shuffled sources & 53.55 & 49.62 & 50.39 & 68.21 & 51.57 \\
Specialist & 58.13 & 57.79 & 60.38 & 68.91 & 55.01 \\
\midrule[0.05pt]
\multicolumn{1}{@{}l}{\raisebox{1.2ex}{\scriptsize \textbf{Score Prediction}}}\\[-6pt]
Shuffled sources & 52.32 & 49.32 & 47.96 & 68.43 & 52.58 \\
Specialist & 56.77 & 55.93 & 56.78 & 68.58 & 56.56 \\
\bottomrule
\end{tabular}
\vspace{-0.5em}
\caption{Comparison of Specialist AutoMQM vs a Specialist Scorer, which predicts float scores only, on the WMT'23 and WMT'24 test sets. For both AutoMQM and Score Prediction, the ``Champion'' setting without filtering of ICL examples is used.\\
\label{tab:metricx_vs_automqm_23_24}
}
\end{table}
\section{Conclusion}
In this work, we have proposed the \textit{Specialist} method for development of automatic evaluation metrics which are specialized to a given test set. We have shown that Specialist AutoMQM dramatically outperforms all existing state-of-the-art span-based machine translation evaluation metrics, on both the WMT'23 and WMT'24 test sets.

Specialist evaluators are easy to implement, as they are simply multi-shot prompted LLMs. Moreover, the Specialist method is task-agnostic, and an immediate avenue for future work would be to apply this method to evaluation of other natural language generation (NLG) tasks. These Specialist metrics could serve as a powerful alternative to human judges in evaluating LLM quality across a wide range of capabilities. Another avenue for future work would be to better understand how to combine ratings from multiple raters, both during creation of ICL examples for Specialist metrics, and for creation of more trustworthy test sets (which are capable of measuring super-human performance). Finally, the Specialist method as framed here requires human-generated ratings to be used as ICL examples, but future work could explore whether LLMs are also capable of generating these ratings.

\clearpage
\section{Limitations}
As shown in \S\ref{sec:raters}, given high human inter-annotator variability (which, as discussed in \S\ref{sec:related-work}, is not an easy problem to resolve, even for expert raters trained in the specific evaluation task), it is important that ratings be collected in a \textit{pseudo-SxS} fashion (i.e., ratings of outputs from the same input should be performed by a fixed rater; see \S\ref{setup:datasets}). This is not yet standard practice for collection of ratings and, to the best of our knowledge, no publicly available, commonly used benchmarks for evaluation of LLM-as-a-Judge models and reward models follow this rating collection procedure. We constructed a Specialist metric for the MT-Bench dataset \citep{zheng2023judging}, which consists of expert-based ratings (pairwise comparisons) of different model outputs for challenging, multi-turn questions. However, our shuffled baseline already outperformed human inter-annotator agreement, and our Specialist metric did not significantly outperform the shuffled baseline. Given that ratings were not collected in a pseudo-SxS fashion, performance improvements beyond the level of human inter-annotator agreement are undefined and cannot be measured.
\bibliography{custom}
\bibliographystyle{icml2024}

\newpage
\appendix
\onecolumn
\section{Implementation Details}
\label{appendix:b}

Figure \ref{fig:automqm_prompt} shows the AutoMQM prompt template, Table \ref{tab:automqm_output} shows an example AutoMQM output, and Figure \ref{fig:da_prompt} shows the direct assessment scoring prompt template.
\begin{figure}[h!]
\begin{tcolorbox}[title=AutoMQM Prompt Template, colback=blue!5!white, colframe=blue!75!black]
\begin{minipage}{\linewidth}
\begin{verbatim}
You are an annotator for the quality of machine translation. Your task is to
identify errors and assess the quality of the translation.
Based on the source segment and machine translation surrounded with triple
backticks, identify error types in the translation and classify them. The
categories of errors are: accuracy (addition, mistranslation, omission,
untranslated text), fluency (character encoding, grammar, inconsistency,
punctuation, register, spelling), style (awkward), terminology (inappropriate
for context, inconsistent use), non-translation, other, or no-error.
Each error is classified as one of three categories: critical, major, and
minor. Critical errors inhibit comprehension of the text. Major errors disrupt
the flow, but what the text is trying to say is still understandable. Minor
errors are technically errors, but do not disrupt the flow or hinder
comprehension.

Make sure your response is a strict and valid json object that could be parsed
with json.loads() in python.

\end{verbatim}

\textbf{ICL examples}
\begin{verbatim}
{source_language} source:
```{source}```
{target_language} translation:
```{translation}```
{errors in JSON format}

{source_language} source:
```{source}```
{target_language} translation:
```{translation}```
{errors in JSON format}

\end{verbatim}

\textbf{Test example}
\begin{verbatim}
{source_language} source:
```{source}```
{target_language} translation:
```{translation}```
\end{verbatim}
\end{minipage}
\end{tcolorbox}
\caption{AutoMQM prompt, with placeholders for \texttt{\{source\_language\}}, \texttt{\{source\}} (for both ICL examples and the test example), \texttt{\{target\_language\}}, \texttt{\{translation\}} (again, for both ICL examples and the test example), and \texttt{\{errors in JSON format\}} (for ICL examples only).}
\label{fig:automqm_prompt} 
\end{figure}

\definecolor{red}{RGB}{255,90,90}
\definecolor{lightred}{RGB}{255,182,182}

\setlength{\tabcolsep}{0.1em}
\begin{CJK*}{UTF8}{gbsn}
\begin{table*}[h!]
\centering
\begin{tabular}{ll}
\toprule
Source & 害得我，从外地驱车200公里赶回来取货！\\
Hypothesis & \hlc[lightred]{I'm sorry that} \hlc[lightred]{we} had to drive 200 kilometers from \hlc[red]{the country} to pick up my goods! \\
\midrule
Output & \small{\texttt{[\{"span": "I'm sorry that", "severity": "minor", "category": "style/unnatural or awkward"\},}} \\
& \small{\texttt{ \{"span": "we", "severity": "minor", "category": "accuracy/mistranslation"\},}} \\
& \small{\texttt{ \{"span": "the country", "severity": "major", "category": "accuracy/mistranslation"\}]}} \\
\bottomrule
\end{tabular}
\caption{Example Specialist AutoMQM output (from the WMT'23 zh$\rightarrow$en test set). As per the AutoMQM prompt (Figure \ref{fig:automqm_prompt}), the output is in JSON format, with fields for error \texttt{span}, \texttt{severity}, and \texttt{category}. The highlighting is added to the hypothesis for illustrative purposes, to indicate the locations of the predicted major (dark red) and minor (light red) errors.}
\label{tab:automqm_output}
\end{table*}
\end{CJK*}

 \begin{figure}[h!]
\begin{tcolorbox}[title=Direct Assessment Scoring Prompt Template, colback=blue!5!white, colframe=blue!75!black]
\begin{minipage}{\linewidth}
\begin{verbatim}
You are a judge for the quality of machine translation. Based on the
source segment, human-generated reference translation, and machine
translation surrounded with triple backticks, your task is to assess
the quality of the machine translation on a continuous scale from 0 to
100. A score of 0 means "No meaning preserved", then the scale goes
through "Some meaning preserved", to "Most meaning preserved and few
grammar mistakes", up to a score of 100, which means "Perfect meaning
nd grammar".
\end{verbatim}

\textbf{ICL examples}
\begin{verbatim}
{source_language} source:
```{source}```
{target_language} translation:
```{translation}```
Score: [[{score}]]

{source_language} source:
```{source}```
{target_language} translation:
```{translation}```
Score: [[{score}]]

\end{verbatim}

\textbf{Test example}
\begin{verbatim}
{source_language} source:
```{source}```
{target_language} translation:
```{translation}```
\end{verbatim}
\end{minipage}
\end{tcolorbox}
\caption{Direct Assessment prompt, with placeholders for \texttt{\{source\_language\}}, \texttt{\{source\}} (for both ICL examples and the test example), \texttt{\{target\_language\}}, \texttt{\{translation\}} (again, for both ICL examples and the test example), and \texttt{\{score\}} (for ICL examples only).}
\label{fig:da_prompt} 
\end{figure}
\section{Supplemental Results}
\label{appendix:a}

\begin{table}[h!]
\centering
\begin{tabular}{l|cc}
\toprule
Number of systems & WMT'23 & WMT'24\\
\midrule
en$\rightarrow$de & 12 & 17 \\
zh$\rightarrow$en & 15 & N/A \\
en$\rightarrow$es & N/A & 13 \\
ja$\rightarrow$zh & N/A & 13 \\
\bottomrule
\end{tabular}
\caption{Number of translation systems for each language pair (WMT'23 and WMT'24)}
\label{tab:num_systems_wmt}
\end{table}

\begin{table}[h!]
\centering
\begin{tabular}{r|cc}
\toprule
Avg \# ICL examples / \\
Avg \# errors per test example & en$\rightarrow$de & zh$\rightarrow$en \\
\midrule
No filtering & 11.0/34.8 & 14.0/30.7 \\
Filtered & 10.5/27.4 & 13.3/25.8 \\
\bottomrule
\end{tabular}
\caption{Average number of ICL examples and average number of total errors in ICL examples, per test example (WMT'23 test set). The filtered setting removes all translations from ICL examples which are exact matches to the test translation, and removes all individual errors from ICL examples which exactly match a ground-truth error span in the test translation.}
\label{tab:error_stats_23}
\end{table}
\begin{table}[h!]
\centering
\begin{tabular}{r|ccc}
\toprule
Avg \# ICL examples / \\
Avg \# errors per test example & en$\rightarrow$de & en$\rightarrow$es & ja$\rightarrow$zh \\
\midrule
No filtering & 17.0/22.8 & 13.0/5.8 & 13.1/13.3 \\
Filtered & 16.0/19.4 & 12.4/5.6 & 12.8/12.8 \\
\bottomrule
\end{tabular}
\caption{Average number of ICL examples and average number of total errors in ICL examples, per test example (WMT'24 test set). The filtered setting removes all translations from ICL examples which are exact matches to the test translation, and removes all individual errors from ICL examples which exactly match a ground-truth error span in the test translation.}
\label{tab:error_stats_24}
\end{table}

\setlength{\tabcolsep}{0.75em}
\begin{table*}[ht]
\centering
\begin{tabular}{lcccccc}
\toprule
& \multicolumn{3}{c}{en$\rightarrow$de} & \multicolumn{3}{c}{zh$\rightarrow$en} \\
\cmidrule{2-4} \cmidrule{5-7}
& F1 & Precision & Recall & F1 & Precision & Recall \\
\midrule
\multicolumn{1}{@{}l}{\raisebox{1.2ex}{\scriptsize \textbf{Baselines}}}\\[-6pt]
1a) XCOMET-XXL-QE & 32.71\phantom{*} & 28.66\phantom{*} & 38.10\phantom{*} & 34.29\phantom{*} & 39.70\phantom{*} & 30.18\phantom{*} \\
1b) GEMBA-MQM-QE & 29.80\phantom{*} & 32.04\phantom{*} & 27.85\phantom{*} & 34.17\phantom{*} & 39.87\phantom{*} & 29.89\phantom{*} \\
1c) Shuffled sources & 31.49\phantom{*} & 28.34\phantom{*} & 35.45\phantom{*} & 37.80$^*$ & 32.79$^*$ & 44.62$^*$ \\
1d) Fixed, different source & 22.85\phantom{*} & 26.08\phantom{*} & 20.35\phantom{*} & 31.27\phantom{*} & 34.54\phantom{*} & 28.58\phantom{*} \\
\midrule
\multicolumn{1}{@{}l}{\raisebox{1.2ex}{\scriptsize \textbf{Specialist}}}\\[-6pt]
2a) Specialist & \textbf{45.71}$^*$ & \textbf{45.04}$^*$ & \textbf{46.40}$^*$ & \textbf{57.47}$^*$ & \textbf{54.05}$^*$ & \textbf{61.36}$^*$ \\
2b) Specialist + Filter & 38.32$^*$ & 39.05$^*$ & 37.61$^*$ & 50.72$^*$ & 49.32$^*$ & 52.21$^*$ \\
\bottomrule
\end{tabular}
\vspace{-0.5em}
\caption{Specialist AutoMQM results on the WMT'23 test set. See \S\ref{sec:models} for a description of all of the Baseline and Specialist systems, and see \S\ref{ablation:filtering} for a description of the filtered setting. Results for the ``Shuffled sources'' and ``Fixed, different source'' baselines are reported as the average over 10 runs with different random seeds. See Table \ref{tab:baselines_avg_variance} for the variance across runs. Asterisks ($*$) indicate scores which are statistically significantly better than XCOMET (row 1a), according to a paired permutation test.
\label{tab:main_table_wmt23}
}
\end{table*}

\setlength{\tabcolsep}{0.5em}
\begin{table*}[ht]
\centering
\begin{tabular}{llcccccc}
\toprule
& & \multicolumn{3}{c}{en$\rightarrow$de} & \multicolumn{3}{c}{zh$\rightarrow$en} \\
\cmidrule{3-5} \cmidrule{6-8}
& & F1 & Precision & Recall & F1 & Precision & Recall \\
\midrule
Shuffled sources & AVG & 31.49 & 28.34 & 35.45 & 37.80 & 32.79 & 44.62 \\
Shuffled sources & STDEV & 0.71 & 0.62 & 1.01 & 0.33 & 0.33 & 0.50 \\
\midrule
Fixed, different source & AVG & 22.85 & 26.08 & 20.35 & 31.27 & 34.54 & 28.58 \\
Fixed, different source & STDEV & 0.76 & 0.93 & 0.97 & 0.48 & 0.57 & 0.67 \\
\bottomrule
\end{tabular}
\caption{Average (AVG) character-level F1, precision, and recall over 10 runs of the ``Shuffled sources'' and ``Fixed, different source'' baselines with different random seeds. Standard deviation (STDEV) over the 10 runs is also reported.
\label{tab:baselines_avg_variance}
}
\end{table*}

\setlength{\tabcolsep}{0.4em}
\begin{table*}[h!]
\centering
\begin{tabular}{lccccccccc}
\toprule
& \multicolumn{3}{c}{en$\rightarrow$de} & \multicolumn{3}{c}{en$\rightarrow$es} & \multicolumn{3}{c}{ja$\rightarrow$zh} \\
\cmidrule{2-4} \cmidrule{5-7} \cmidrule{8-10}
& F1 & Precision & Recall & F1 & Precision & Recall & F1 & Precision & Recall \\
\midrule
\multicolumn{1}{@{}l}{\raisebox{1.2ex}{\scriptsize \textbf{Baselines}}}\\[-6pt]
1a) XCOMET-XXL-QE & 24.28\phantom{*} & 19.63\phantom{*} & 31.83\phantom{*} & 10.11\phantom{*} & 6.02\phantom{*} & 31.42\phantom{*} & 14.30\phantom{*} & 11.80\phantom{*} & 18.16\phantom{*} \\
1b) Shuffled sources & 26.12\phantom{*} & 19.67\phantom{*} & 38.84\phantom{*} & 26.12$^*$ & 19.67$^*$ & 38.84$^*$ & 26.44$^*$ & 32.46$^*$ & 22.30$^*$ \\
1c) Fixed, different source & 18.23\phantom{*} & 19.82\phantom{*} & 16.87\phantom{*} & 14.89\phantom{*} & 11.09\phantom{*} & 22.65\phantom{*} & 26.77$^*$ & 32.25$^*$ & 22.89$^*$ \\
\midrule
\multicolumn{1}{@{}l}{\raisebox{1.2ex}{\scriptsize \textbf{Specialist}}}\\[-6pt]
2a) Specialist & \textbf{43.04}$^*$ & \textbf{39.16}$^*$ & \textbf{47.76}$^*$ & \textbf{26.58}$^*$ & \textbf{20.05}$^*$ & \textbf{39.43}$^*$ & \textbf{37.16}$^*$ & \textbf{38.06}$^*$ & \textbf{36.30}$^*$ \\
2b) Specialist + Filter & 32.83$^*$ & 31.07$^*$ & 34.79$^*$ & 25.58$^*$ & 19.34$^*$ & 37.79$^*$ & 35.73$^*$ & 36.83$^*$ & 34.69$^*$ \\
\bottomrule
\end{tabular}
\vspace{-0.5em}
\caption{Specialist AutoMQM results on the WMT'24 test set. Note that en$\rightarrow$es ratings were \textit{not} collected in a \textit{pseudo-SxS} fashion (see \S\ref{setup:datasets}), which explains the smaller performance delta between the Specialist method and the baselines for this language pair. See \S\ref{sec:models} for a description of all of the Baseline and Specialist systems, and see \S\ref{ablation:filtering} for a description of the filtered setting. Asterisks ($*$) indicate scores which are statistically significantly better than XCOMET (row 1a), according to a paired permutation test. Note: The en$\rightarrow$es MQM data was not collected in a pseudo-SxS fashion, so ratings from different raters were presented as ICL examples in the Specialist setup for this language pair.
\label{tab:main_table_wmt24}
}
\end{table*}

\definecolor{lightred}{RGB}{255,106,106}
\definecolor{lightgreen}{RGB}{144,238,144}

\begin{CJK*}{UTF8}{gbsn}
\begin{table*}[h!]

    \centering

    \begin{tabular}{lp{\exampleWidth}}
    \toprule
    Source & 34CM的床垫不是一般的厚，不要床直接睡床垫都可以了。\\
    \midrule
    Test Example Hypothesis & A 34CM mattress \hlc[lightred]{is not usually thick}, \hlc[lightgreen]{so it is not necessary to place the bed directly on the mattress}. \\
    \midrule
    ICL Examples & A 34cm mattress is not \hlc[lightred]{typically} thick, \hlc[lightred]{you} could even sleep directly on the mattress without a bed. \\
    & \hlc[lightred]{The 34CM mattress is not generally thick}, and you can sleep directly on the mattress without a bed. \\
    & The 34CM mattress is unusually thick, \hlc[lightred]{you} can sleep directly on the mattress without a bed. \\
    & \hlc[lightred]{34CM mattress is not generally thick, do not sleep directly on the mattress can be.} \\
    & The 34cm mattress \hlc[lightred]{is not usually thick}, so you can sleep directly on the mattress without the bed. \\
    & \hlc[lightred]{34 cm mattress is not as thick as usual}, but the beds can be used directly. \\
    & The 34CM mattress \hlc[lightred]{is not so thick}, you can just sleep on the mattress without the bed. \\
    & \dots \\
    \bottomrule
    \end{tabular}

    \vspace*{1em}

    \begin{tabular}{lp{\exampleWidth}}
    \toprule
    Source & 吓得我把收藏夹里的其乐都删了。\\
    \midrule
    Test Example Hypothesis & I was so scared that I deleted all the \hlc[lightgreen]{games} from my favorites. \\
    ICL Example Errors & I was so scared that I deleted all the \hlc[lightred]{music} in my favorites. \\
    & I was so scared that I deleted all the \hlc[lightred]{fun} in my favorites. \\
    & I'm \hlc[lightred]{afraid} I deleted all the \hlc[lightred]{music} from my collection. \\
    & \hlc[lightred]{I was in the middle of a conversation.} \\
    & I am so scared \hlc[lightred]{to} remove all the \hlc[lightred]{items} in my collection. \\
    & \hlc[lightred]{Scared} me so \hlc[lightred]{much I} deleted all my favorites of \hlc[lightred]{its music}. \\
    & \dots \\
    \bottomrule
    \end{tabular}
    \caption{Examples of where Specialist AutoMQM predicts errors not present in ICL examples (WMT'23 zh$\rightarrow$en test set). Green highlighting in the Test Example Hypothesis shows where Specialist AutoMQM correctly predicted an error span that was \textit{not} present in the ICL examples, while red highlighting indicates a span (correctly) copied from ICL examples. Red highlighting in the ICL Examples indicates the error spans that were marked by human MQM annotators (and provided to Specialist AutoMQM as demonstrations).}
    \label{tab:new_error_pred_examples}
\end{table*}
\end{CJK*}

\definecolor{lightred}{RGB}{255,106,106}
\definecolor{lightgreen}{RGB}{144,238,144}

\begin{CJK*}{UTF8}{gbsn}
\begin{table*}[h!]

    \centering

    \begin{tabular}{lp{\exampleWidth}}
    \toprule
    Source & 标题上还是顾客,正文中就变成客户了。 \\
    \midrule
    ICL Example Hypothesis & The title is still a \hlc[lightred]{customer}, but the text becomes a customer. \\
    \midrule
    Test Example Hypothesis & The title still refers to the \hlc[lightgreen]{customer}, but in the body of the text, it has changed to client. \\
    \bottomrule
    \end{tabular}

    \vspace*{1em}

    \begin{tabular}{lp{\exampleWidth}}
    \toprule
    Source & 让一个身上3处伤口的老人下床开门收快递还要找零钱付费！\\
    \midrule
    ICL Example Hypothesis & Let an old man with 3 wounds get out of bed and open the door to receive the courier and \hlc[lightred]{change to pay}! \\
    \midrule
    Test Example Hypothesis & To get an old man with 3 wounds on his body to get out of bed and open the door to receive the package, he still has to find \hlc[lightgreen]{change to pay}! \\
    \bottomrule
    \end{tabular}
    \caption{Examples of where Specialist AutoMQM correctly abstains from copying errors in ICL examples (WMT'23 zh$\rightarrow$en test set). Red highlighting indicates that the span was marked as an error by the human MQM annotators, and green highlighting indicates that the span was not marked as an error.}
    \label{tab:copy_abs_examples}
\end{table*}
\end{CJK*}

\begin{table*}
\centering
\begin{tabular}{l|cc|cc|c}
\toprule
& \multicolumn{2}{c}{Round1 ICL Examples} & \multicolumn{2}{c}{Round2 ICL examples} & \\
\midrule
test\_set\_rater\_id & icl\_rater\_id & F1 & icl\_rater\_id & F1 & num\_examples \\
\midrule
rater1 & rater1 & 0.39 & rater8 & 0.30 & 672 \\
rater2 & rater2 & 0.50 & rater6 & 0.35 & 540 \\
rater3 & rater3 & 0.45 & rater1 & 0.33 & 564 \\
rater4 & rater4 & 0.46 & rater10 & 0.33 & 552 \\
rater5 & rater5 & 0.46 & rater7 & 0.26 & 588 \\
rater6 & rater6 & 0.51 & rater4 & 0.36 & 540 \\
rater7 & rater7 & 0.46 & rater9 & 0.29 & 492 \\
rater8 & rater8 & 0.37 & rater2 & 0.27 & 528 \\
rater9 & rater9 & 0.48 & rater3 & 0.34 & 528 \\
rater10 & rater10 & 0.51 & rater5 & 0.26 & 516 \\
\bottomrule
\end{tabular}
\caption{Specialist AutoMQM performance on WMT'23 en$\rightarrow$de Round1, when prompting using Round1 vs Round2 ICL examples, broken out by rater split. In each round of WMT'23 ratings, there are a total of 10 en$\rightarrow$de raters. The examples in the test set are then approximately split evenly across all raters (such that all translations of the same source segment are allocated to the same rater). Note that the variance across raters when using different-rater (Round2) ICL examples is not very high, and using Round1 ICL examples outperforms Round2 ICL examples for every split}
\label{tab:cross_rater_f1}
\end{table*}

\subsection{Additional Ablations}
\subsubsection{Filtering ICL Examples to Remove Exact-Match Errors}
\label{ablation:filtering}

Here, we isolate the effect on performance of showing Specialist AutoMQM errors in ICL examples which are an exact match to a ground-truth error in the test translation. In particular, we filter ICL examples to i) remove errors with the same span (but not necessarily the same category or severity) as ground-truth errors present in the test translation, and ii) entirely exclude all translations (rather than just removing exact-match errors) in ICL examples which are exact matches to the test translation.

As expected, filtering the ICL examples by removing all error spans present in the ground truth (``Specialist + Filter'' setting, row 2b in Tables \ref{tab:main_table_wmt23} and \ref{tab:main_table_wmt24} for WMT'23 and WMT'24, respectively) does incur some degradation in performance relative to the ``Specialist'', but still significantly outperforms all baselines, including the state-of-the-art XCOMET and ``Shuffled source'' models. Also note that filtering to remove individual errors from ICL examples in some sense unfairly disadvantages the model, since this procedure excludes real errors from the demonstrations, and these are, in fact, exactly those errors which would be correct for the model to predict.

For this filtered Specialist AutoMQM, we computed exact match rates with respect to i) ground truth errors spans in the test translations and ii) error spans present in ICL examples. The results are shown in Table \ref{tab:icl_ground_truth_exact_match}. Observe that, even though the model was not shown any of the ground truth errors in the provided demonstrations, 17.1\% (for en$\rightarrow$de) and 23.7\% (for zh$\rightarrow$en) of the errors that it predicts are exact matches to the ground truth, while 26-27\% of the errors that it predicts are exact matches to ICL example errors spans. If predicted error spans which are either sub-spans or super-spans of errors present in ICL examples are also counted as matches, then the match rate more than doubles, to 65-68\%. This suggests that Specialist AutoMQM is also taking into account the semantics of the errors in the ICL examples, and is able to generalize its predictions to account for modified versions of these errors present in the test translations.
\begin{table}
\centering
\begin{tabular}{l|cc}
\toprule
Exact Match Error \% & en$\rightarrow$de & zh$\rightarrow$en \\
\midrule
1) Ground Truth & 17.10 & 22.70 \\
2) ICL Examples & 27.52 & 26.10 \\
3) ICL Examples \\
(incl. sub-span + super-span) & 65.25 & 68.79 \\
\bottomrule
\end{tabular}
\caption{Exact match error rate of Specialist AutoMQM predictions, as a percentage of total predicted errors, with respect to the ground truth error spans (row 1) and error spans present in ICL examples (row 2). Row 3 shows the match rate when predicted error spans which are either sub-spans or super-spans of errors present in ICL examples are also counted as matches. Results are presented for ``Champion + Filtered'' WMT'23 en$\rightarrow$de Specialist AutoMQM (Table \ref{tab:main_table_wmt23_wmt24}, row 2b), so the model is not shown demonstrations of any errors with exact match to the ground truth errors in the test translation.}
\label{tab:icl_ground_truth_exact_match}
\end{table}

\subsubsection{Can We Do Better? Augmenting Specialist AutoMQM with More ICL Examples}

Specialist AutoMQM only uses the same-source ratings from the test set as ICL examples, which limits the number of ICL examples to num\_systems - 1. Modern LLMs can handle much longer context than these examples occupy, so in this ablation, we investigate whether augmenting the same-source ICL examples provided to Specialist AutoMQM with other ICL examples from the test set can further enhance performance. In particular, for each test set example, we first provide the ICL examples from the shuffled baseline, then concatenate the ICL examples from Specialist AutoMQM. As shown in Table \ref{tab:additional_icl_examples}, augmenting Specialist AutoMQM with additional examples results in a small \textit{drop} in character-level F1, due to lower recall (despite a small improvement in precision). Recall that in Figure \ref{fig:icl_examples_scaling}, we also saw that Specialized AutoMQM performance saturates at around 10 (same-source) ICL examples. This suggests that there is not substantial headroom to improve AutoMQM's performance by filling up the LLM's long context window, either with same-source or difference-source ICL examples. 
\setlength{\tabcolsep}{0.5em}
\begin{table*}[ht]
\centering
\begin{tabular}{lcccccc}
\toprule
& \multicolumn{3}{c}{en$\rightarrow$de} & \multicolumn{3}{c}{zh$\rightarrow$en} \\
\cmidrule{2-4} \cmidrule{5-7}
& F1 & Precision & Recall & F1 & Precision & Recall \\
\midrule
Shuffled sources & 31.12 & 27.93 & 35.13 & 37.62 & 32.45 & 44.74 \\
Specialist AutoMQM & \textbf{45.71} & 45.04 & \textbf{46.40} & \textbf{57.47} & 54.05 & \textbf{61.36} \\
Shuffled Sources + Specialist AutoMQM & 44.75 & \textbf{46.17} & 43.42 & 57.36 & \textbf{55.79} & 59.02 \\
\bottomrule
\end{tabular}
\vspace{-0.5em}
\caption{Comparison of prompting with i) only shuffled sources, ii) only same-source examples, or iii) both. Adding additional ICL examples gives higher precision at the cost of lower recall.
\label{tab:additional_icl_examples}
}
\end{table*}

\end{document}